\documentclass[a4paper,fleqn]{cas-sc}

\usepackage[numbers]{natbib}
\usepackage{amssymb, amsfonts}
\usepackage{multicol}
\usepackage{multirow}
\usepackage{algorithm}
\usepackage{algorithmic}
\usepackage{comment}

\usepackage{pifont}
\definecolor{bronze}{rgb}{1.0, 0.49, 0.0}

\usepackage{array}

\usepackage[caption=false,font=footnotesize]{subfig}

\usepackage[utf8]{inputenc}
\usepackage{tabularray}

\newcommand{\leo}[1]{{{#1}}}

\newcommand{\silvio}[1]{{\textcolor{red}{}}}

\begin{document}
\let\WriteBookmarks\relax
\def\floatpagepagefraction{1}
\def\textpagefraction{.001}

\shorttitle{}    

\shortauthors{}  

\title [mode = title]{Flyweight FLIM Networks for Salient Object Detection in Biomedical Images}  

\tnotemark[1] 

\tnotetext[1]{} 

%

\author[1]{Leonardo M. João}[orcid=0000-0003-4625-7840]

\cormark[1]


\ead{leomelo168@gmail.com}


\credit{Conceptualization, Investigation, Methodology, Software, Writing – original draft}

\affiliation[1]{organization={Instituto de Computação, Universidade Estadual de Campinas (UNICAMP)},
            addressline={Rua Saturnino de Brito, 2}, 
            city={Campinas},
            postcode={13083-872}, 
            state={Sao Paulo},
            country={Brazil}}

\author[2]{Jancarlo F. Gomes}

\credit{Data curation and validation}
\affiliation[2]{organization={Faculdade de Ciências Médicas, Universidade Estadual de Campinas (UNICAMP},
            addressline={Rua Saturnino de Brito, 2}, 
            city={Campinas},
            postcode={13083-872}, 
            state={Sao Paulo},
            country={Brazil}}

\author[3]{Silvio J. F. Guimaraes}

\credit{Conceptualization, Supervision, Writing – review and editing}
\affiliation[3]{organization={Institute of Computing Sciences, Pontifical Catholic University of Minas Gerais},
            addressline={Rua Saturnino de Brito, 2}, 
            city={Belo Horizonte},
            postcode={30535-901}, 
            state={Minas Gerais},
            country={Brazil}}
            
\author[4]{Ewa Kijak}

\credit{Conceptualization, Supervision, Writing – review and editing}
\affiliation[4]{organization={University of Rennes, IRISA, Inria},
            addressline={Rua Saturnino de Brito, 2}, 
            city={Rennes},
            postcode={35000}, 
            state={},
            country={France}}
            
\author[1]{Alexandre X. Falcao}
\credit{Conceptualization, Supervision, Writing – review and editing}
\cortext[1]{Corresponding author}

\fntext[1]{}


\begin{abstract}
Salient Object Detection (SOD) with deep learning often requires substantial computational resources and large annotated datasets, making it impractical for resource-constrained applications. Lightweight models address computational demands but typically strive in complex and scarce labeled-data scenarios. Feature Learning from Image Markers (FLIM) learns an encoder's convolutional kernels among image patches extracted from discriminative regions marked on a few representative images, dismissing large annotated datasets, pretraining, and backpropagation. Such a methodology exploits information redundancy commonly found in biomedical image applications. This study presents methods to learn dilated-separable convolutional kernels and multi-dilation layers without backpropagation for FLIM networks. It also proposes a novel network simplification method to reduce kernel redundancy and encoder size. By combining a FLIM encoder with an adaptive decoder, a concept recently introduced to estimate a pointwise convolution per image, this study presents very efficient (named flyweight) SOD models for biomedical images. Experimental results in challenging datasets demonstrate superior efficiency and effectiveness to lightweight models. By requiring significantly fewer parameters and floating-point operations, the results show competitive effectiveness to heavyweight models. These advances highlight the potential of FLIM networks for data-limited and resource-constrained applications with information redundancy.
\end{abstract}



\begin{keywords}
Lightweight networks, \sep salient object detection, \sep image feature learning, \sep fully convolutional neural networks.
\end{keywords}

\maketitle
 
\section{Introduction}\label{sec:intro}

Salient Object Detection (SOD) focuses on highlighting objects that stand out in an image \cite{borji2019salient}. Advances in SOD methods have benefited various computer vision tasks, such as image retrieval \cite{al2021saliency} and image compression \cite{wang2021focus}. State-of-the-art SOD methods predominantly leverage deep learning to develop saliency models that outperform traditional heuristic-based approaches \cite{borji2019salient}. Deep learning is particularly effective for creating general-purpose models that can be specialized for different applications. Although this generic-to-specific model adaptation strategy can yield impressive results, it becomes challenging or infeasible in important scenarios -- e.g., model learning in biomedical image applications with complex and scarce labeled data and model deployment in low-powered computers and embedded devices. Moreover, energy-intensive models raise serious environmental concerns~\cite{schwartz2020green}.

Lightweight Convolutional Neural Networks (CNNs) have been proposed for SOD~\cite{gao2020highly,liu2021samnet,lin2022lightweight,liang2023meanet}, offering reduced computational costs while achieving substantial improvements over traditional SOD methods. These networks feature fewer parameters and employ efficient operations -- e.g., the decomposition of the standard convolution into depthwise and pointwise convolutions in MobileNet~\cite{sandler2018mobilenetv2} and the use of multiple dilation rates with the same separable convolutional kernels in SAMNET \cite{liu2021samnet}. These methods have shown competitive results to deep models, particularly on more specific problems. However, training lightweight models remains challenging in complex and scarce labeled-data scenarios.

\begin{figure}[!t]
    \centering
    \includegraphics[width=1.0\linewidth]{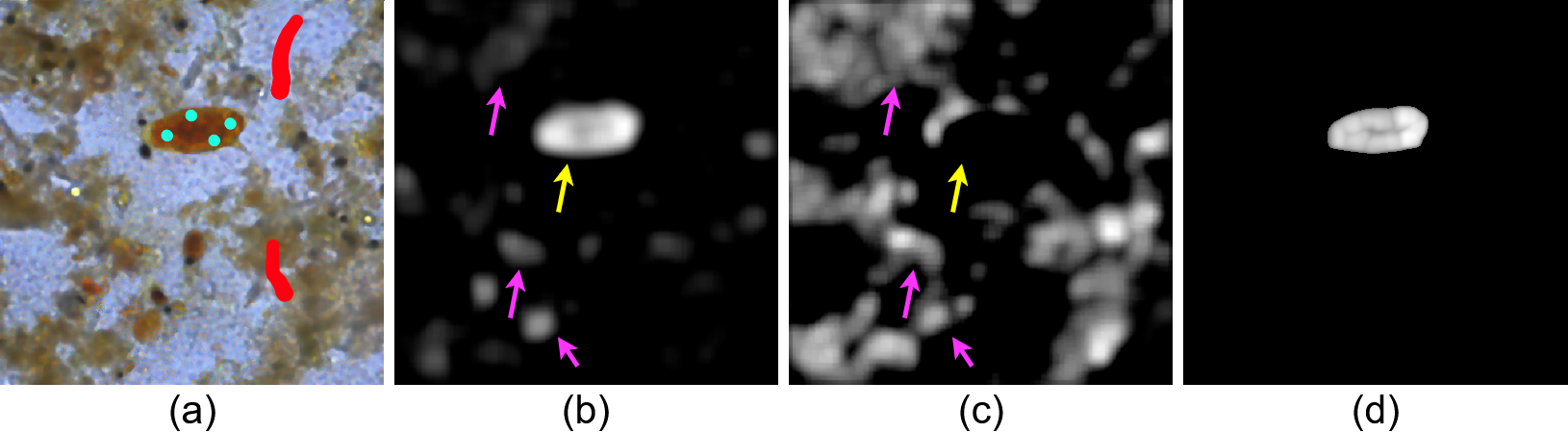}
    \caption{(a) Original Image with background and object components. User-drawn markers are shown in cyan (object) and red (background); (b) Foreground activation channel with the object (yellow arrow) and some false positives (pink arrows) activated; (c) Background activation channel, in which the object (yellow arrow) is not activated;  (d) Resulting saliency map from an adaptive decoder.}
   \label{fig:adaptive-decoder-example}
\end{figure}

Alternatives to the generic-to-specific training paradigm are required for complex and scarce labeled-data scenarios. Feature Learning from Image Markers (FLIM) is a recent methodology \cite{de2020feature} under investigation, in which the convolutional kernels of an encoder are discovered among image patches from discriminative regions of a few representative images. In FLIM, an expert can directly identify regions with local visual patterns (image patches) that distinguish classes (or objects) by drawing markers, which dismisses backpropagation to identify the attention (relevant) regions in an image. Figure~\ref{fig:adaptive-decoder-example}a, for example, illustrates cyan (disks) on a parasite egg (object) and red (scribbles) markers on background regions for SOD. Convolution can be interpreted as a similarity function \cite{JoaoCBF24}, where a positive activation at a given pixel suggests a strong resemblance between the image patch centered at that pixel and the visual pattern of a kernel. The kernels of each encoder's layer are estimated as cluster centers in a patch dataset extracted from all marker pixels using the input features of that layer. For SOD, the output of any given layer should mainly present foreground (Figure~\ref{fig:adaptive-decoder-example}b) or background (Figure~\ref{fig:adaptive-decoder-example}c) activation channels, although uncertain activation channels may occur. Given the similarities between the object and some background parts, false positives may be observed in foreground activation channels but tend to reduce along with the layers. By combining a FLIM encoder with an adaptive decoder, a concept recently introduced to estimate a pointwise convolutional kernel for each input image, one can design a SOD model without backpropagation~\cite{joao2023flyweight}. One can define different types of adaptive decoders~\cite{soares2024adaptive} and improve SOD by adding object delineation strategies~\cite{salvagnini2024improving}. Such decoders rely on a predetermined rule to classify the output channel from any layer as foreground (positive weight), background (negative weight), or uncertain (zero weight). The term ``adaptive'' stems from the on-the-fly weight estimation for each input image. Through pointwise convolution followed by activation, background activations are subtracted from foreground activations, emphasizing the object while reducing or eliminating false positives (Figure \ref{fig:adaptive-decoder-example}d).

\begin{figure}[b!]
    \centering
         \includegraphics[width=0.6\linewidth]{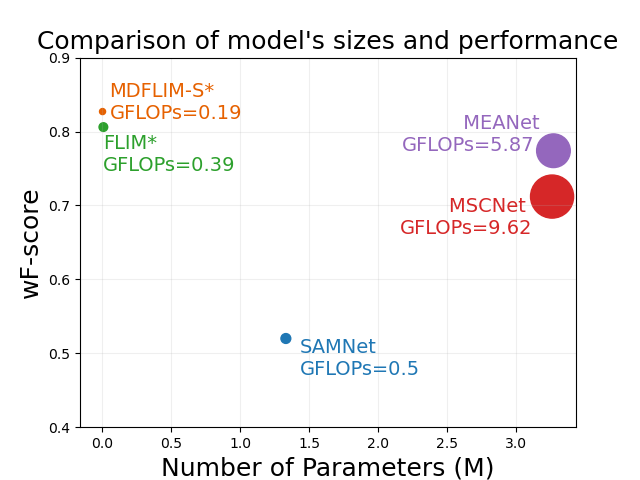} 
      \caption{Comparison of model's size and performance among FLIM and  baselines from the state-of-the-art for SOD on the \textit{S. mansoni} eggs dataset. The size of the circles represents the number of Giga Floating Point Operations (GFLOPS).}
    \label{fig:comparison_graphic}
\end{figure}

The FLIM methodology substitutes human effort in annotating large image datasets with expert intervention in marker drawing and representative image selection~\cite{cerqueira2024selection} from an unlabeled dataset. The methodology may not be suitable for applications with many classes (objects) in heterogeneous (e.g., natural) images,  unless labeled data are available with alternatives for automated identification of discriminative regions in representative images. In this study, we assume information redundancy for SOD in biomedical image applications. The study incorporates new methods in the FLIM framework to learn lightweight operations: dilated-separable convolutional kernels and multi-dilation layers without backpropagation. Dilated-separable kernels are depthwise separable kernels with multiple dilation factors. Given $m$ regular kernels with $f$ channels each, the depthwise decomposition replaces them by the convolution with the mean kernel (a depthwise result) and $m$ pointwise convolutions that average the $f$ channels of the depthwise result with different weights. Such weights have been learned by backpropagation~\cite{sandler2018mobilenetv2}. This study introduces a new method to learn those weights from the image markers. The regular FLIM kernels in each given layer are separated into their mean kernel, and statistics from the activation channels of the depthwise result are used to estimate the weights of the pointwise convolutions. We can create multi-dilation layers in regular and separable convolutions using multiple dilation rates in the FLIM kernel estimation method. The method incorporates multi-scale information in the adaptive decoder, generating an efficient and effective SOD model.   

To further reduce network size and computational load, we propose a method for automatically removing redundant kernels in FLIM Networks. When multiple kernels have similar coefficients, they collectively emphasize a single feature, which a single representative kernel with an increased magnitude could do instead. In standard CNNs, such simplification is challenging due to strong interdependencies among pretrained weights, exemplified by layer-collapse~\cite{tanaka2020pruning}. However, FLIM learns kernels layer by layer, making kernel removal less disruptive to the overall model. Thus, we iteratively remove redundant kernels based on a uniqueness score while enhancing the magnitude of the most similar remaining kernel. By applying this simplification at the end of each layer during training, we can achieve significantly smaller CNNs with comparable performance. Since these models are orders of magnitude smaller than conventional lightweight CNNs -- which typically range between 1M-5M (million) parameters -- we refer to them as flyweight, with model sizes often below 100K (thousand) parameters.

We use two challenging biomedical datasets for validation: one for detecting \textit{Schistosoma mansoni} eggs in microscopy images and another for detecting brain tumors in MRI. Our FLIM models require significantly fewer operations while achieving superior results to standard FLIM networks (baselines). We show that the performance of our CNNs is competitive with state-of-the-art heavyweight models and superior to lightweight models in scenarios with scarce labeled data. To illustrate the efficiency and performance gains, Figure \ref{fig:comparison_graphic} depicts the efficiency and performance metrics of the proposed and lightweight networks. Heavyweight methods were not included in this figure because their efficiency metrics are in a different order of magnitude, invalidating proper analysis of the differences between ours and lightweight models.

As main contributions, this paper presents (i) a method for learning dilated-separable convolutional kernels from image markers; (ii) a method to learn multi-scale features from image markers and incorporate them in each FLIM layer and the adaptive decoder; and (iii) a network simplification technique that leverages unique characteristics of FLIM to build flyweight SOD networks.

This manuscript is organized as follows: In Section~\ref{sec:rw}, heavy and lightweight SOD methods with their proposed improvements are described in a overview of related works from the literature. Later, in Section~\ref{sec:background}, definitions and mathematical interpretations required for a good understanding of the proposed solutions are presented together with an overview of FLIM encoder training and Adaptive Decoder application. In Section~\ref{sec:method} we present our proposed methodology for learning flyweight CNNs with all its components. Then, the experimental setup, results and discussions are presented in Section~\ref{sec:experiments}. Lastly, conclusions are drawn in Section~\ref{sec:conclusion}.

\section{Related Work}\label{sec:rw}
  Traditional SOD methods rely on handcrafted features but struggle with the diversity and complexity of images \cite{borji2019salient}. Deep learning approaches, particularly fully convolutional (encoder-decoder) networks, have enhanced robustness and performance by learning directly from data~\cite{wang2021salient}. Such models first encode features into deep and low-resolution representations, which are then upscaled by a symmetrical decoder~\cite{ronneberger2015u}. Further enhancements include reformulated dropout and advanced upsampling modules~\cite{zhang2017learning}, as well as pyramid pooling and global guidance for improved feature aggregation~\cite{liu2019simple}. To leverage multiple feature scales, methods such as DGRL \cite{wang2018detect} and BASNet \cite{qin2019basnet} employ coarse-to-fine enhancement modules to achieve high-resolution saliency. BASNet produces high-quality saliency maps using a hybrid boundary-aware loss and a secondary encoder-decoder network. More recently, to fully utilize multiscale information, U²Net \cite{qin2020u2} introduced a nested U-Net architecture, maintaining efficiency with low computational requirements despite a high parameter count. This approach achieves state-of-the-art results without the need for pretraining. Overall, multiscale information has been instrumental in advancing the performance of SOD models.
   
Despite the extensive literature, a recent survey~\cite{zhou2024benchmarking} benchmarked several methods under a unified experimental setup, finding that BASNet achieved the best overall performance across multiple datasets (notably, U²Net was not evaluated). Therefore, in this paper, we use as reference points for heavyweight networks BASNet as the best performing SOD method, and U²Net for it showed no need for pretraining. 

    Recent efforts have focused on reducing the computational cost of SOD models. Gao \textit{et al.} \cite{gao2020highly} introduced a lightweight CNN using cross-stage fusion to leverage multiscale information efficiently. HVPNet \cite{liu2020lightweight} incorporates hierarchical visual perception modules to capture multiscale contexts effectively, while SAMNet \cite{liu2021samnet} applies dilated separable convolutions to enhance model efficiency.  Other approaches enhance lightweight models by using MobileNetV2 \cite{sandler2018mobilenetv2} as a backbone. MSCNet \cite{lin2022lightweight} employs a Multiscale Context Extraction module and an Attention-based Pyramid Feature Aggregation mechanism to leverage multiscale features effectively. MEANet \cite{liang2023meanet} improves saliency map resolution for optical remote-sensing images by integrating a Multiscale Edge-embedded Attention Module with a Multilevel Semantic Guidance Module, achieving state-of-the-art results and surpassing heavier CNNs in remote sensing tasks. Although these methods are lightweight, they still require substantial annotated data for (pre)training, and most need a few billion floating-point operations for execution. We have selected MEANet, MSCNet, and SAMNet as baselines representing lightweight models from the state of the art.
    \begin{figure}[t!]
    \centering
         \includegraphics[width=0.8\textwidth]{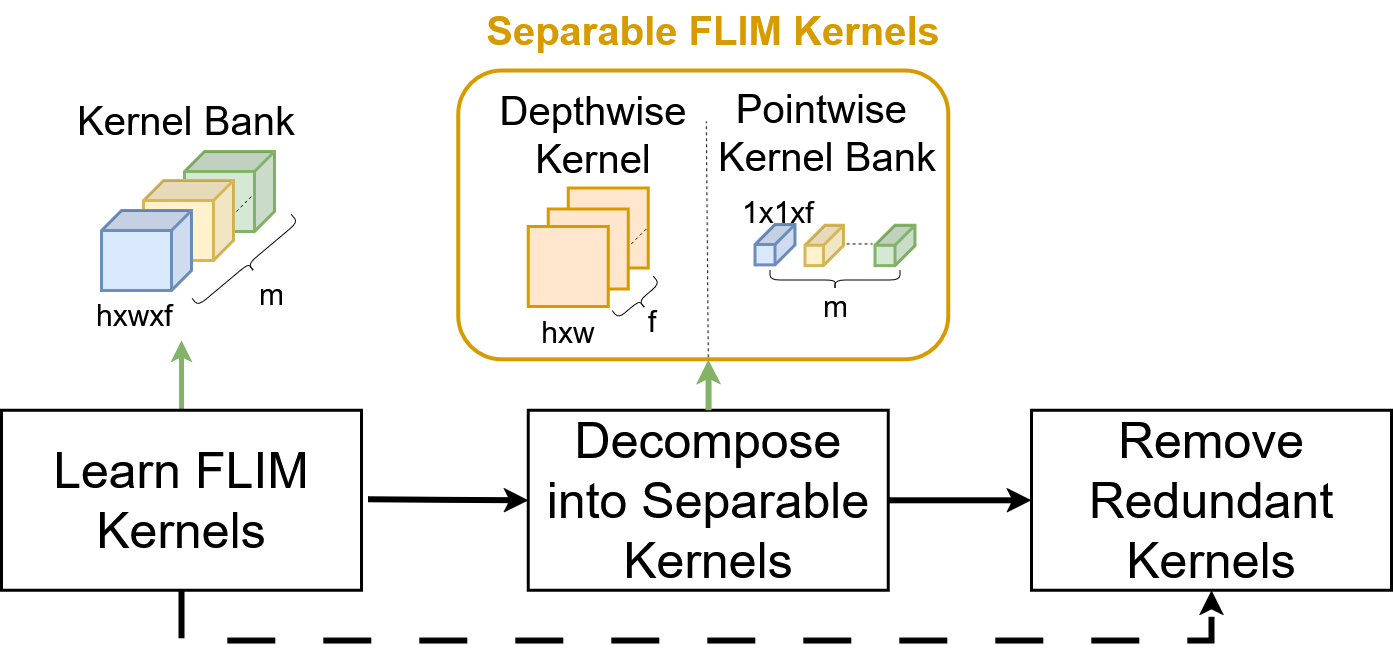} 
     \caption{Steps for learning a flyweight FLIM layer are shown. Green arrows illustrate the output of each layer, while black arrows indicate possible data flows, with dotted lines representing paths used for training a simplified, non-separable CNN.}
    \label{fig:method_diagram}
\end{figure}

\section{Background}\label{sec:background}

This section presents basic definitions with the formalism needed to explain the proposed methods, the main steps involved in the design of a FLIM encoder, and the adaptive decoder used for this study. 

\noindent \textbf{Images and Image Patches:} Let $\mathbf{X} \in \mathbb{R}^{h \times w \times f}$ represent an image, where $h \times w$ are its spatial dimensions, and $f$ denotes the number of channels. For any $i \in [1, h]$ and $j \in [1, w]$, the feature vector of the pixel at position $p=(i, j)$ is $\mathbf{x}_{ij} \in \mathbb{R}^f$, with $x_{ijb} \in \mathbb{R}$ representing its $b$-th feature, where $b \in [1, f]$.

Generally, $\mathcal{A}_d = \bigcup\limits_{x = -\lfloor \frac{a}{2} \rfloor}^{\lfloor \frac{a}{2} \rfloor}{\bigcup\limits_{y = \lfloor \frac{a}{2} \rfloor}^{\lfloor \frac{a}{2} \rfloor}{\{(d\times x, d\times y)\}}}$ defines a set of displacements of size $|\mathcal{A}_d| = a \times a$, where $a$ is an odd integer and $d \in \mathbb{N}$ is a dilation factor. A non-dilated adjacency can be represented as $\mathcal{A}_1$. Thus, $\mathcal{A}_d(p) = \{(i-d\times x, j-d\times y) : (x,y) \in \mathcal{A}_d \}$ defines the adjacency of a pixel $p$.

Lastly, an image patch $\mathbf{p}_{p} \in \mathbb{R}^{a \times a \times f}$ is a sub-image that includes all $f$ features for each of the $a \times a$ pixels within the neighborhood defined by $\mathcal{A}_d(p)$. Padding is applied to ensure that each pixel in the image has a valid patch, even near the borders.


\noindent \textbf{Kernels and Convolutions:} A kernel $\mathbf{k} \in \mathbb{R}^{a \times a \times f}$ is a matrix with the same dimensions (shape) as an image patch. A standard/regular convolution of an image with a kernel produces an output image $\mathbf{Y} \in \mathbb{R}^{h \times w \times 1}$, such that: 

\begin{equation}\label{eq:pixelconv}
    y_{ij} = \sum\limits_{x = 1}^{a}{\sum\limits_{y = 1}^{a}{\sum\limits_{b = 1}^{f}{q_{xyb}\times k_{xyb}}}},
\end{equation}
\noindent with $q_{xyb} \in \mathbf{p}_{p}$, $k_{xyb}\in \mathbf{k}$, $i\in [1,h], j\in [1,w]$, and $y_{ij} \in \mathbf{Y}$.

For computational efficiency in lightweight CNNs, depthwise separable convolution is often used \cite{sandler2018mobilenetv2}. This operation splits a standard convolution into a depthwise convolution followed by a pointwise convolution, significantly reducing computational costs. It involves a depthwise kernel $\mathbf{k'} \in \mathbb{R}^{a \times a \times f}$ and  pointwise kernels $\mathbf{k^\star} \in \mathbb{R}^{1 \times 1 \times f}$. For a CNN layer with $m$ standard kernels, using one depthwise kernel along with $m$ pointwise kernels provides a close approximation to the regular convolution.


First, the depthwise convolution performs a separate convolution for each channel (i.e., a single convolution with the mean kernel of the given kernel bank), producing an output image $\mathbf{Y}' \in \mathbb{R}^{h \times w \times f}$, computed as:
\begin{equation}\label{eq:depthwiseconv}
    y'_{ijb} = \sum\limits_{x = 1}^{a}{\sum\limits_{y = 1}^{a}{q_{xyb}\cdot k'_{xyb}}},
\end{equation}
\noindent with $y'_{ijb} \in \mathbf{Y'}$, $q_{xyb} \in \mathbf{p}_{p}$, $k'_{xyb}\in \mathbf{k'}$, $i\in [1,h], j\in [1,w]$, and $b \in [1, f]$. Finally, the pointwise convolutions combine the channels with different weights, yielding $\mathbf{Y} \in \mathbb{R}^{h \times w \times 1}$, defined as:
\begin{equation}\label{eq:pointwiseconv}
    y_{ij} = \sum\limits_{b = 1}^{f}{y'_{ijb}\cdot k^{\star}_{ijb}},
\end{equation}

\noindent in which $k^{\star}_{ijb}\in \mathbf{k^{\star}}$.

A convolutional layer can consist of $m$ kernels (or the mean kernel and $m$ pointwise kernels in the case of depthwise separable convolutions) and additional operations such as pooling and ReLU activation. To account for possible stride in the convolution or pooling layers, the output of a convolutional layer is an image $\mathbf{L} \in \mathbb{R}^{h' \times w' \times m}$, where $1 \leq h' \leq h$ and $1 \leq w' \leq w$.

\noindent \textbf{FLIM Encoders:} 
FLIM learns the encoder's kernels from user-drawn markers on discriminative regions of representative images. It interprets the convolution operation as a kernel-patch similarity~\cite{JoaoCBF24}; thus, by selecting representative image patches (cluster centers) as convolutional kernels, the convolutions are expected to activate the discriminative visual patterns in different channels. The encoder learning process in FLIM involves five main steps.

    \begin{enumerate}
      \item \textbf{Image selection:} assuming information redundancy, selecting a few representative images should be sufficient for FLIM. For this work, we selected five images for each dataset by visual inspection.
      
      \item \textbf{Marker drawing:} discriminative regions may be identified using user-drawn scribbles (\textit{e.g.}, Figure \ref{fig:adaptive-decoder-example}). In this work, the designer freely drew scribbles over all object instances and distinct background regions;
    
      \item \textbf{Data preparation:} this step involves applying marker-based normalization~\cite{JoaoCBF24} in the patch dataset derived from (rescaled) markers using the input features for the current layer;
      
      \item \textbf{Kernel Estimation:} given the specified hyperparameters of a convolutional layer, candidate kernels are extracted from patches centered at marked pixels as described in Section \ref{sec:method-flim-kernels}
    
      \item \textbf{Layer execution:} the layer is executed to obtain new image features. The markers are projected onto the input of the next layer and the process loops back to Step (3) when additional layers need to be learned.
\end{enumerate}   

      \leo{\textbf{Adaptive Decoders:}} FLIM encoders can be combined with various predictor types for image classification~\cite{de2020learning, sousa2021cnn}, segmentation~\cite{de2020feature, cerqueira2023building}, and object detection~\cite{JoaoCBF24, joao2023flyweight}. This paper utilizes adaptive decoders~\cite{joao2023flyweight}, whose weights are estimated on the fly for each input image, to generate saliency maps. Such adaptive decoders are typically implemented as pointwise convolutions followed by ReLU activation, with dynamic kernel weight estimation facilitated by an adaptation function. The adaptation function determines whether an activation map should be treated as foreground (positive weight), background (negative weight), or discarded (zero weight). The adaptation function can be application-dependent. For instance, it can expect the foreground to occupy a smaller portion of the image than the background. In this case, the mean activation value of each activation channel is evaluated to determine if it is sufficiently low or high to represent either a foreground or a background channel, respectively. More details and a formal introduction to such decoders are provided in~\cite{soares2024adaptive}.
      

      The decoder is defined as $\mathbf{S} = \text{ReLU}(\mathbf{L} \star \boldsymbol{\alpha})$, where $\boldsymbol{\alpha} \in \mathbb{R}^{1\times 1\times m}$ is a pointwise kernel with $\alpha_b \in \{-1, 0, 1\}$, $b \in [1, m]$, $\mathbf{L} \in \mathbb{R}^{h'\times w'\times m}$ is the output of a layer, and $h'$, $w'$, and $m$ represent the image dimensions and the number of features, respectively. In this work, the adaptation function that defines the weights $\boldsymbol{\alpha}$ is based on the approach proposed in~\cite{JoaoCBF24}, with a slight modification. As described in~\cite{JoaoCBF24}, all weights are initially set to one, and an adaptive function maps them to either a positive, negative or zero value, formally, $\mathbf{H}: \alpha_b \rightarrow \{-\alpha_b, 0, \alpha_b\}$. The adaptation function then set the weights as:

      \begin{equation}
          \mathbf{H}(\alpha_b) = 
        \begin{cases}
            +\alpha,& \text{if } \mu_{b} \leq \tau - \sigma^2 \text{and } \psi_b > 0.1  \\
            -\alpha, & \text{if } \mu_{b}\geq \tau + \sigma^2 \text{and } \psi_b < 0.2\\
            0, & \text{otherwise.}
        \end{cases}
      \end{equation}

      \noindent in which $\{\mu_1, \mu_2, ..., \mu_m\}$ define a distribution of the mean activation of each kernel for the processing image, $\tau$ is the Otsu threshold of said distribution, $\sigma$ its standard deviation, and $\psi_b = \frac{1}{h'w'} \sum\limits_{i = 1}^{h'} \sum\limits_{j = 1}^{w'} t_{ijb}$ represent the foreground ratio of a channel-wise binarization of $\mathbf{L}$ using the Otsu threshold, represented as $\mathbf{T} \in \mathbb{R}^{h'\times w'\times m}$, with $t_{ijb} \in \mathbf{T}$.
      
      The addition of  $\psi$ helps to further eliminate unreliable kernels by discarding any channel with a foreground ratio that does not match its class.

\section{Flyweight FLIM networks with Simplified Dilated Separable Layers}\label{sec:method}
    This section presents the main contributions of this work. It explains how to learn regular FLIM kernels (Section \ref{sec:method-flim-kernels}), how to factorize them into separable ones (Section \ref{sec:method-sep-conv}), how to create multi-dilation layers (Section \ref{sec:method-multi-dilation}), and how to simplify these layers by removing redundant kernels (Section \ref{sec:method-simplification}). Figure \ref{fig:method_diagram} illustrates the steps for learning a flyweight CNN with simplified separable layers. The same network can be used with or without multi-dilation layers, as discussed in Section \ref{sec:method-multi-dilation}.

\subsection{\leo{Learning regular FLIM kernels}}\label{sec:method-flim-kernels}

\noindent FLIM utilizes patches centered on marked pixels using the input features of the current layer. The number of marked patches in the resulting patch dataset is initially balanced to $m_r$ representatives per marker. These patches are further reduced to the layer's convolutional kernels' target number ($m$).

    Formally, $\mathcal{M}$ represents a superset of markers, where each marker $M$ consists of a set of connected pixels. Candidate kernels are generated by considering all patches centered on each pixel within a marker $M$, forming $\mathcal{P}_M = \bigcup\limits_{\forall p \in M}{\mathbf{p}_{p}}$. Given the variability in marker sizes, the representative kernel sets $\mathcal{K}_M \subset \mathcal{P}_M$ are derived from each marker using k-means clustering, with the resulting $m_r$ cluster centers representing each marker. The union of these sets across all markers yields $\mathcal{K}_U = \bigcup\limits_{\forall M \in \mathcal{M}}{\mathcal{K}_M}$.

    A final k-means clustering is applied to $\mathcal{K}_U$, yielding $m$ cluster centers that serve as the regular convolutional kernels, thus forming the final kernel bank $\mathcal{K} = \{\mathbf{k}_1, \mathbf{k}_2, \dots, \mathbf{k}_{m}\}$, where each $\mathbf{k}_c \in \mathbb{R}^{a\times a\times f}$ for $c \in [1, m]$, such that $\mathcal{K} \subset \mathcal{K}_U$.
    
    Note that applying a dilated adjacency to the patches makes the resulting kernel suitable for dilated convolutions.

\subsection{Decomposition into separable kernels}\label{sec:method-sep-conv}
\noindent Given $m$ kernels with $f$ channels in a kernel bank, they are separated into the mean kernel with $f$ channels and $m$ pointwise kernels that average the $f$ channels with different weights.

\begin{figure}[]
    \centering
         \includegraphics[width=0.8\textwidth]{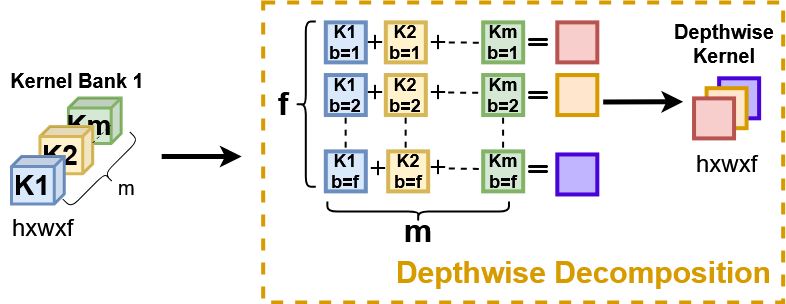} 
     \caption{Diagram of kernel factorization into depthwise separable ones.}
    \label{fig:depthwise-diagram}
\end{figure}

\noindent\textbf{Depthwise:} The depthwise kernel is the mean kernel of the given kernel bank $\mathcal{K}$, \textit{i.e.}, each channel of the mean kernel is obtained by the average of the  corresponding coefficients from  the kernel bank (Figure \ref{fig:depthwise-diagram}). This kernel is then used in a depthwise convolution, resulting in an output image with  $f$ channels. Formally, $\mathbf{k}' = \frac{1}{m} \sum\limits_{\forall \mathbf{k}_c \in \mathcal{K}}{\mathbf{k}_c}$.

\begin{figure}[b!]
    \centering
         \includegraphics[width=0.7\textwidth]{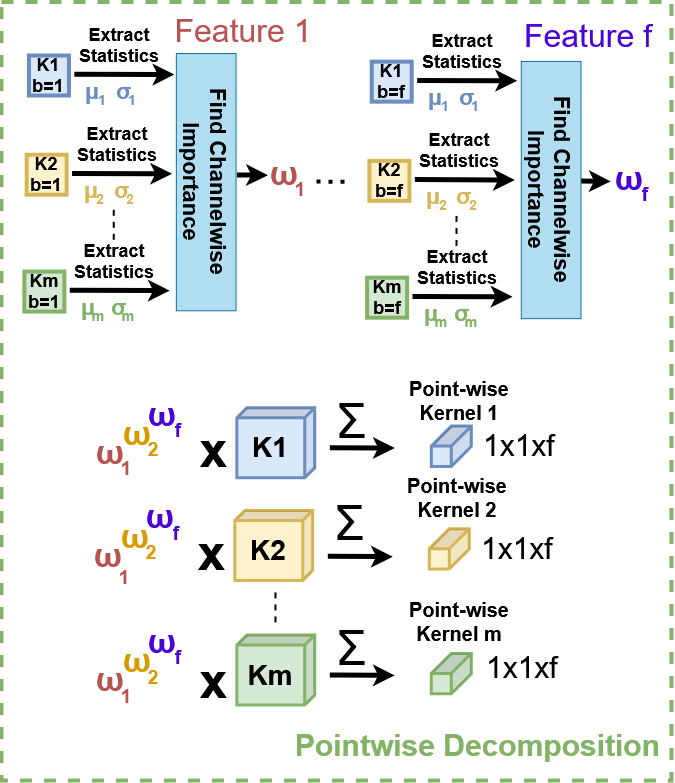} 
     \caption{Diagram of kernel factorization into pointwise separable ones.}
    \label{fig:pointwise-diagram}
\end{figure}

    \noindent\textbf{Pointwise:} We derive a pointwise kernel bank $\mathcal{K}^\star = \{\mathbf{k}^\star_1, \mathbf{k}^\star_2, ..., \mathbf{k}^\star_{m}\}$ from $\mathcal{K}$ by analyzing each kernel's coefficients, the bank's channelwise mean, and the standard deviation for each input channel $b \in [1,f]$. We start by computing a channel-wise importance $\omega_b$ and subsequently use it in a weighted sum of coefficients to define an importance value for each kernel at each channel. The pointwise filter is then defined as a vector containing all channelwise weighted sums for each kernel (Figure \ref{fig:pointwise-diagram}).

    Formally a pointwise kernel is derived from a regular one, such that $\boldsymbol{\phi} : \mathcal{K} \rightarrow \mathcal{K}^\star$, such that for each kernel $\mathbf{k_c}$ with $c \in [1,m~]$, $\boldsymbol{\phi}(\mathbf{k_c}) = [\phi_1, \phi_2, ..., \phi_f]$, and $\phi_b = \frac{\omega_b}{a^2}\sum\limits_{i = 1}^{a}{\sum\limits_{j = 1}^{a}{k_{c_{ij}}}}$, where $b \in [1, f]$. The channel importance is defined as $\omega_b = \frac{1}{\beta} \mu_b \sigma_b$, where $\beta = \sum\limits_{b = 1}^{f}{\mu_b}$ is a normalization factor computed over the mean coefficients and standard deviations of each channel, represented by $\mu_b = \frac{1}{a^2\cdot m}\sum\limits_{c = 1}^{m}{\sum^{a}\limits_{i = 1}{\sum^{a}\limits_{j = 1}{k_{c_{ij}}}}}$ and $\sigma_b = \frac{1}{a^2\cdot m}\sum\limits_{c = 1}^{m}{\sum^{a}\limits_{i = 1}{\sum^{a}\limits_{j = 1}{(k_{c_{ij}} - \mu_b)^2}}}$, respectivelly.

\subsection{Learning multi-dilation FLIM layers}\label{sec:method-multi-dilation}
\noindent     
    Following the separable dilation convolution~\cite{liu2021samnet}, multiple dilated convolutions are computed with varying dilation factors $\mathcal{D}$, and their results are combined through a sum. The kernel coefficients for each dilation factor stay the same. The multiple dilated convolution follows Equation \ref{eq:pixelconv}, but extends it, such that:
\begin{equation}\label{eq:dilatedconv}
    y_{ij} = \sum\limits_{d = 1}^{|\mathcal{D}|}{\sum\limits_{x = 1}^{a}{\sum\limits_{y = 1}^{a}{\sum\limits_{b = 1}^{f}{q^d_{xyb}\cdot k_{xyb}}}}},
\end{equation}
    \noindent where $q_{xyb} \in \mathbf{p}^d_{ij}$, and $\mathbf{p}^d_{(i,j)} \in \mathbf{X}$ represents a dilated image patch with dilation factor $d \in \mathbb{N}$.
    
    Similarly, dilated separable convolutions are achieved by adding the summation to Equations \ref{eq:depthwiseconv} and \ref{eq:pointwiseconv}.
\subsection{Simplification of FLIM networks}\label{sec:method-simplification}
    Even though, in FLIM Networks, kernels are extracted from marked regions, some redundancy is expected due to multiple markers being placed on similar regions or multiple kernels being extracted from each marker, particularly in homogeneous regions.

    To identify redundant kernels, we start by computing a uniqueness score $\boldsymbol{\Upsilon} : \mathbf{k} \rightarrow \mathbb{R}$ , defined by:
    \begin{equation}
        \boldsymbol{\Upsilon}(\mathbf{k}_i) = \frac{1}{m}\sum\limits_{j = 1}^{m}{\textbf{D}^\textbf{2}(\mathbf{k}_i, \mathbf{k}_j)}.
    \end{equation}

    \noindent where $\textbf{D}^\textbf{2} : \mathbf{k} \times \mathbf{k} \rightarrow \mathbb{R}$ is the squared distance between two kernels, defined as $\textbf{D}^\textbf{2}(\mathbf{k}_i, \mathbf{k}_j) = \sum\limits_{x = 1}^{a}{\sum\limits_{y = 1}^{a}{\sum\limits_{b = 1}^{f}{(\mathbf{k}_{ixyb} - \mathbf{k}_{jxyb})^2}}}$. 

    With the uniqueness scores computed, let $\mu_\Upsilon$ denote the average uniqueness of the kernel bank. We define a set of removable kernels $\mathcal{R} \subset \mathcal{K}$ such that $\Upsilon(\mathbf{k}i) < \mu_\Upsilon \implies \mathbf{k}_i \in \mathcal{R}$. Lastly, let $\mathbf{N} : \mathbf{k}_i \rightarrow \mathbf{k}_j$ map a kernel to its closest neighbor, such that:

    \begin{equation}
        \mathbf{N}(\mathbf{k}_i) = \arg \min_{k_j \in \mathcal{K}, \, k_i \neq k_j} \textbf{D}^\textbf{2}(k_i, k_j).
    \end{equation}

    With these definitions, each kernel is removed one at a time if it is part of the removable set and its nearest neighbor has not already been removed. After removal, one of the remaining kernels will represent the removed kernel, and its magnitude will increase. The amount in which the magnitude of the representative kernels increase is tied to the number of removed kernels it represents. Therefore, when a kernel is removed, we increase the number of kernels its closest neighbor represents by one. Lastly, we multiply the magnitude of each kernel by the number of kernels it represents, increasing their magnitude (one if only representing itself, and increasing by one for each closest neighbor removed). This process is repeated until every removable kernel has been processed. To iteratively simplify the network, the process can be repeated $n$ times at each layer. The procedure is described in Algorithm \ref{alg:simplification}.
    
\begin{algorithm}
\caption{Layer Simplification Algorithm}
\label{alg:simplification}
\begin{algorithmic}[1]
\STATE \textbf{Prerequisites:}
\STATE \hspace{0.5cm} A kernel bank $\mathcal{K}$
\STATE \hspace{0.5cm} Number of iterations $n$
\STATE \hspace{0.5cm} Vector $\mathbf{c}$ of ones indicating the number of kernels each kernel represents 

\hspace{-0.5cm}\hrulefill
\STATE \textbf{Repeat} $n$ times:
\STATE \hspace{0.5cm} Compute all uniqueness scores $\boldsymbol{\Upsilon}$
\STATE \hspace{0.5cm} Compute all nearest neighbors \textbf{N}
\STATE \hspace{0.5cm} Compute the mean uniqueness $\mu_\Upsilon$
\STATE \hspace{0.5cm} \textbf{For} each $\mathbf{k}_i$ in $\mathcal{K}$:
\STATE \hspace{1.0cm} \textbf{If} $\boldsymbol{\Upsilon}(\mathbf{k}_i) < \mu_{\Upsilon}$ \textbf{then}
\STATE \hspace{1.5cm} Add $\mathbf{k}_i$ to $\mathcal{R}$
\STATE \hspace{0.5cm} \textbf{For} each $\mathbf{k}_i$ in $\mathcal{R}$:
\STATE \hspace{1.0cm} \textbf{If} \textbf{N}($\mathbf{k}_i$) $\in \mathcal{K}$ \textbf{then}
\STATE \hspace{1.5cm} nn $\leftarrow$ \textbf{N}($\textbf{k}_i$) /* Find closest neighbor */
\STATE \hspace{1.5cm} $\mathbf{c}_{nn} \leftarrow \mathbf{c}_{nn}+1$ /* Increase number of kernels nn represents */
\STATE \hspace{1.5cm} \textbf{Remove}($\textbf{k}_i, \mathcal{K}$)
\STATE \hspace{1.5cm} \textbf{Remove}($c_i$, $\mathbf{c}$)
\STATE \hspace{0.5cm} \textbf{IncreaseMagnitudes}($\mathcal{K}$, $\mathbf{c}$)
\end{algorithmic}
\end{algorithm}

The simplification can be applied to fully trained FLIM Networks; however, once a layer $l$ has been simplified, all subsequent layers must be re-trained.

\section{Experimental results and discussion}\label{sec:experiments}
    This section presents the experimental setup (Section \ref{subsec:setup}), with split selection, hyperparameter configuration, compared methods (FLIM network types), datasets, post-processing steps, and evaluation criteria. Next, various FLIM models are compared  (\ref{subsec:comparison-flim}), including a regular model (baseline) and its variants with multi-dilation layers, separable convolutions, and network simplification. Finally, to provide context, FLIM is compared to other state-of-the-art methods on the presented datasets (\ref{subsec:comparison-sota}), along with a discussion of its limitations and the need for post-processing in the case of the \textit{S. mansoni eggs} dataset.
    
\subsection{Experimental setup}\label{subsec:setup}
\noindent\textbf{Splits and cross validation:} The datasets underwent a random 70-30 split for creating one set of training/validation and one set of testing, respectively. From the 70\% subset, three training splits were created with five manually selected images each, with the remaining images being used for validation. Visual analysis guided the image selection to represent object variability in each training set adequately. The same splits were used for all methods, \textit{i.e.}, and the same five images were used for training (backpropagation-based methods were also pre-trained on SOD datasets).

\noindent\textbf{\textit{S. mansoni} eggs dataset:} The \textit{S. mansoni} eggs (\textbf{\textit{S. mansoni eggs}}) dataset is public\cite{soares2024adaptive}\footnote{Available at: \url{https://github.com/LIDS-Datasets/schistossoma-eggs}} and comprises 1219 images, each with dimensions of $400 \times 400 \times 3$ pixels. Pixel-wise annotations are available. The images do not always contain an object of interest, often feature cluttered backgrounds, and sometimes include fecal impurities that occlude the eggs.

\noindent\textbf{Brain tumor dataset:} The brain tumor (\textbf{Tumor}) dataset is a modified version of the BraTS 2021 training dataset, which is part of the most prominent brain tumor segmentation challenge. The original dataset comprises 1251 samples of volumetric images. However, this modification includes three slices of $240 \times 240 \times 1$ pixels from each FLAIR image, along with the corresponding ground truth, represented as binary rather than multiple segmentation classes.


\noindent\textbf{FLIM hyperparameters:} The network architectures are presented in Table \ref{tab:parameters}, where $k$ represents the kernel size, $d$ the dilation ratio (for FLIM and FLIM-S), $m$ the number of kernels, and maxp and avgp indicate either max or average pooling along with the pooling size, while $s$ denotes the pooling stride. All layers utilize marker-based normalization and ReLU activation. For the \textit{S. mansoni} eggs dataset, the number of kernels per marker is $k_m=5$, and for the Tumor dataset, $k_m=4$. In multi-dilation layers, we empirically set $\mathcal{D} = {1,2,3}$, consistent with SAMNet's values \cite{liu2021samnet}. All hyperparameters were defined by the network designer during model training.

\noindent\textbf{Other methods hyperparameters:} We used the default (author defined) hyper-parameters for all methods, apart from the learning-rate and number of epochs, which were empirically set on the lightweight models due to a lack of convergence when using the default values. On the lightweight models, we also changed the hard-coded Imagenet normalization values (mean and standard deviation) with the respective mean and standard-deviation for each dataset used. Without the normalization values changes the methods could not converge.

\noindent\textbf{Simplification parameters:} The simplification parameters were optimized using a grid search, where training was performed on a predefined network and training setup with a fixed architecture (apart from the reduction in the number of kernels) and image markers. The results for different parameters and their impact are presented and discussed in Sections \ref{sec:results-flim} and \ref{sec:discuss-flim}.

 \begin{table}
     \centering
     \resizebox{0.95\linewidth}{!}{%
\begin{tblr}{
  colspec = {X[c,m]X[c,m]X[c,m]X[c,m]X[c,m]},
  rowsep = 1pt,
  hlines = {0.5pt},
  vlines = {0.5pt}, 
}
  Dataset & Layer 1 & Layer 2 & Layer 3 & Layer 4     \\
  \textit{S. mansoni eggs} & $k = 3$, $d = 1$ \newline $m = 32$\newline maxp$=5$, s$=1$& $k = 3$, $d = 1$ \newline $m = 32$\newline avgp$=5$, s$=1$& $k = 3$, $d = 1$ \newline $m = 8$\newline maxp=7, s$=1$& $k = 3$, $d = 7$ \newline $m = 8$\newline maxp=5, s$=1$ \\
  Tumor & $k = 3$, $d = 1$ \newline $m = 16$\newline maxp$=3$, s$=2$& $k = 3$, $d = 1$ \newline $m = 32$\newline maxp=3, s$=2$& $k = 3$, $d = 1$ \newline $m = 64$\newline maxp=3, s$=2$& ---
\end{tblr}
          }
     \caption{\leo{FLIM Network architectures.}}
     \label{tab:parameters}
 \end{table}

\noindent\textbf{FLIM network types}: Results are reported for FLIM (baseline) and FLIM-S (with separable convolutions), with multiple dilations per layer denoted by MD (MDFLIM and the one with dilated-separable convolutions, MDFLIM-S). Simplified models take their network type followed by a \textbf{*}, such as FLIM*.

\noindent\textbf{Post-processing step:} As demonstrated in \cite{soares2024adaptive}, post-processing is crucial for the \textit{S. mansoni} eggs dataset due to the high number of false positives observed for all methods, FLIM-based and SOTA models. We present results with and without post-processing, along with discussions of its utility. A size filter is applied to remove small and large connected components, and a graph-based segmentation algorithm~\cite{bragantini2018graph} is used to improve delineation. The component size thresholds are individually optimized for each method using a grid search on the validation set. In the graph-based segmentation process, object seeds are estimated from an eroded saliency map using adjacency radius 1, while background seeds are estimated from its dilated version using adjacency radius 30 (fast dilation is computed based on the Euclidean distance transform). 

\noindent\textbf{Evaluation criteria:} We employed two performance metrics and two efficiency metrics. For performance, we used the weighted F-measure ($F^\omega_\beta$) \cite{margolin2014evaluate} and the Mean Absolute Error (MAE). The statistical significance of the results was assessed using the Wilcoxon test~\cite{wilcoxon1992individual}. For efficiency, we considered the number of parameters (\#Params) and floating-point operations (FLOPs (G)).

\noindent\textbf{Computer setup:} All FLIM-related experiments were conducted on a personal computer equipped with an NVIDIA RTX 3060ti GPU with 8 GB of VRAM and a 12th Gen Intel(R) Core(TM) i7-12700K processor. The deep-learning methods were executed on a server featuring four NVIDIA RTX A6000 GPU and an Intel(R) Xeon(R) Gold 5220R processor.

\subsection{Comparison among FLIM models}\label{subsec:comparison-flim}
    In Section \ref{sec:results-flim}, we begin by presenting the quantitative results for the test set across all flyweight model variations, including separable, multi-dilation, and simplified models. Next, we provide the mean and standard deviation results for the different validation splits. Finally, in Section \ref{sec:discuss-flim}, we present qualitative results, discuss the results and offer conclusions and explanations.

\subsubsection{Results}\label{sec:results-flim}
For all tables, the best results for each metric are in bold, and the best overall model's row is highlighted in green.

Out of the three training-validation splits, the models resulting from the split with the best validation performance were selected for evaluation on the test set, and their results are presented in Table~\ref{tab:test-flim-results}.  The table also shows the best achievable simplification for each model, \textit{i.e.}, the model with the highest reduction percentage while maintaining a maximum decrease of $0.01$ in $F^\omega_\beta$.

\begin{table}[!h]
 \centering
    \resizebox{0.7\linewidth}{!}{%
    \begin{tabular}{l|c|c|c|c}
    \hline
    \rowcolor{gray!75}\textcolor{blue}{\textbf{\textit{S. mansoni Eggs}}}
                    &\#Params    &FLOPs(G)   & $F^\omega_\beta$   & MAE\\ \hline 
    FLIM            &12.73(K)  &0.69&0.820&0.705\\ \hline
    FLIM*           &7.06(K)   &0.39&0.820&0.822\\ \hline
    \textbf{MDFLIM} &12.73(K)  &2.08&0.809&0.965\\ \hline
    \textbf{MDFLIM}*&6.39(K)   &0.61&0.826&0.609\\ \hline
    \textbf{FLIM-S} &2.22(K) &0.11 &0.787&0.764\\ \hline
    \textbf{FLIM-S}* &\textbf{1.01(K)} &\textbf{0.05} &0.801&0.817\\ \hline
    \rowcolor{green!35}
    \textbf{MDFLIM-S} &2.06(K) &0.34 &0.817&0.639\\ \hline
    \rowcolor{green!35}
    \textbf{MDFLIM-S}*&1.15(K) &0.19 &\textbf{0.837}&\textbf{0.484}\\ \hline
    \hline 
    \rowcolor{gray!75}
    \textcolor{blue}{\textbf{Tumor}}
                    &\#Params    &FLOPs(G)   & $F^\omega_\beta$   & MAE\\ \hline 
    FLIM            &41.06(K)   &0.58 &0.706&2.301\\ \hline
    FLIM*           &8.12(K)   &0.21 &0.723&2.447\\ \hline
    \textbf{MDFLIM} &41.06(K)   &1.74 &0.671&3.487\\ \hline
    \textbf{MDFLIM}*&8.11(K)   &0.27 &0.722&2.416\\ \hline
    \textbf{FLIM-S} &5.81(K)&0.08 &0.680&2.544\\ \hline
    \textbf{FLIM-S}*&\textbf{2.44(K)} &\textbf{0.03} &0.703&2.813\\ \hline
    \rowcolor{green!35}
    \textbf{MDFLIM-S} &5.81(K) &0.24 &0.731&\textbf{1.864}\\ \hline
    \rowcolor{green!35}
    \textbf{MDFLIM-S}*&3.13(K) &0.14 &\textbf{0.739}&2.469\\ \hline
    \end{tabular}
    }
    \caption{Test-set quantitative results for the models with best validation performance.}
    \label{tab:test-flim-results}
\end{table}

The results indicate that the addition of multi-dilation layers did not improve regular FLIM networks on either dataset. However, multi-dilation separable convolutions demonstrated significant performance gains compared to single-scale counterparts. As for the simplified models, they showed a notable increase in efficiency, often accompanied by a recurrent performance improvement compared to their non-simplified versions, except for regular FLIM on the \textit{S. mansoni Eggs} dataset.

When considering the reduction in the number of parameters and GFLOPs, the decrease was significant when comparing separable models to regular ones, with more than 10x fewer parameters and approximately 6x fewer operations for both datasets. For the simplified models, similar results were observed, achieving about half the number of parameters and operations.

To assess the statistical significance of the differences in results, Table \ref{tab:statistic-significance} presents a comparison between each pair of network types using the Wilcoxon test. Most results are statistically significant, except for some simplified models (\textcolor{bronze}{\textbf{x}}), which exhibit statistically similar performance results with improved efficiency.

\begin{table}[!h]
 \centering
    \resizebox{\linewidth}{!}{%
    \begin{tabular}{l|c|c|c|c|c|c|c|c}
    \hline
    \rowcolor{gray!75}\textcolor{blue}{\textbf{\textit{S. mansoni Eggs}}} 
                    & FLIM & FLIM* & MDFLIM & MDFLIM* & FLIM-S & FLIM-S* & MDFLIM-S & MDFLIM-S*\\ \hline 
    FLIM    & \textcolor{bronze}{\textbf{x}} & \textcolor{bronze}{\textbf{x}} & \ding{51} & \ding{51} & \ding{51} & \ding{51} & \ding{51} &  \ding{51} \\ \hline
    FLIM*   & \textcolor{bronze}{\textbf{x}} & \textcolor{bronze}{\textbf{x}}& \ding{51} & \ding{51} & \ding{51} & \ding{51} & \ding{51} &  \ding{51} \\ \hline
    MDFLIM & \ding{51} & \ding{51} & \textcolor{bronze}{\textbf{x}} & \ding{51} & \ding{51} & \ding{51} & \ding{51} &  \ding{51} \\ \hline
    MDFLIM*& \ding{51} & \ding{51} & \ding{51} & \textcolor{bronze}{\textbf{x}} & \ding{51} & \ding{51} & \ding{51} &  \ding{51} \\ \hline
    FLIM-S & \ding{51} & \ding{51} & \ding{51} & \ding{51} & \textcolor{bronze}{\textbf{x}} & \textcolor{bronze}{\textbf{x}} & \ding{51} &  \ding{51} \\ \hline
    FLIM-S* & \ding{51} & \ding{51} & \ding{51} & \ding{51} & \textcolor{bronze}{\textbf{x}} & \textcolor{bronze}{\textbf{x}} & \ding{51} &  \ding{51} \\ \hline
    MDFLIM-S & \ding{51} & \ding{51} & \ding{51} & \ding{51} & \ding{51} & \ding{51} & \textcolor{bronze}{\textbf{x}}&  \textcolor{bronze}{\textbf{x}} \\ \hline
    MDFLIM-S*& \ding{51} & \ding{51} & \ding{51} & \ding{51} & \ding{51} & \ding{51} & \textcolor{bronze}{\textbf{x}} &  \textcolor{bronze}{\textbf{x}} \\ \hline
    \hline 
    \rowcolor{gray!75}
    \textcolor{blue}{\textbf{Tumor}}
                    & FLIM & FLIM* & MDFLIM & MDFLIM* & FLIM-S & FLIM-S* & MDFLIM-S & MDFLIM-S*\\ \hline 
    FLIM   & \textcolor{bronze}{\textbf{x}} & \ding{51} & \ding{51} & \ding{51} & \ding{51} & \ding{51} & \ding{51} &  \ding{51} \\ \hline
    FLIM*  & \ding{51} & \textcolor{bronze}{\textbf{x}} & \ding{51} & \ding{51} & \ding{51} & \ding{51} & \ding{51} &  \ding{51} \\ \hline
    MDFLIM & \ding{51} & \ding{51} & \textcolor{bronze}{\textbf{x}} & \ding{51} & \ding{51} & \ding{51} & \ding{51} &  \ding{51} \\ \hline
    MDFLIM*& \ding{51} & \ding{51} & \ding{51} & \textcolor{bronze}{\textbf{x}} & \ding{51} & \ding{51} & \ding{51} &  \ding{51} \\ \hline
    FLIM-S & \ding{51} & \ding{51} & \ding{51} & \ding{51} & \textcolor{bronze}{\textbf{x}} & \ding{51} & \ding{51} &  \ding{51} \\ \hline
    FLIM-S* & \ding{51} & \ding{51} & \ding{51} & \ding{51} & \ding{51} & \textcolor{bronze}{\textbf{x}} & \ding{51} &  \ding{51} \\ \hline
    MDFLIM-S & \ding{51} & \ding{51} & \ding{51} & \ding{51} & \ding{51} & \ding{51} & \textcolor{bronze}{\textbf{x}} &  \textcolor{bronze}{\textbf{x}} \\ \hline
    MDFLIM-S*& \ding{51} & \ding{51} & \ding{51} & \ding{51} & \ding{51} & \ding{51} & \textcolor{bronze}{\textbf{x}} &  \textcolor{bronze}{\textbf{x}} \\ \hline
    \hline 
    \end{tabular}
    }
    \caption{Statistical analysis of different models considering the Wilcoxon test on $F^\omega_\beta$ and MAE with a threshold of $0.05$. For each cell,  \ding{51} and \textbf{x} indicates whether or not there is statistical significance between the results, respectively.}
    \label{tab:statistic-significance}
\end{table}

\begin{figure}[h]
    \centering
         \includegraphics[width=0.8\textwidth]{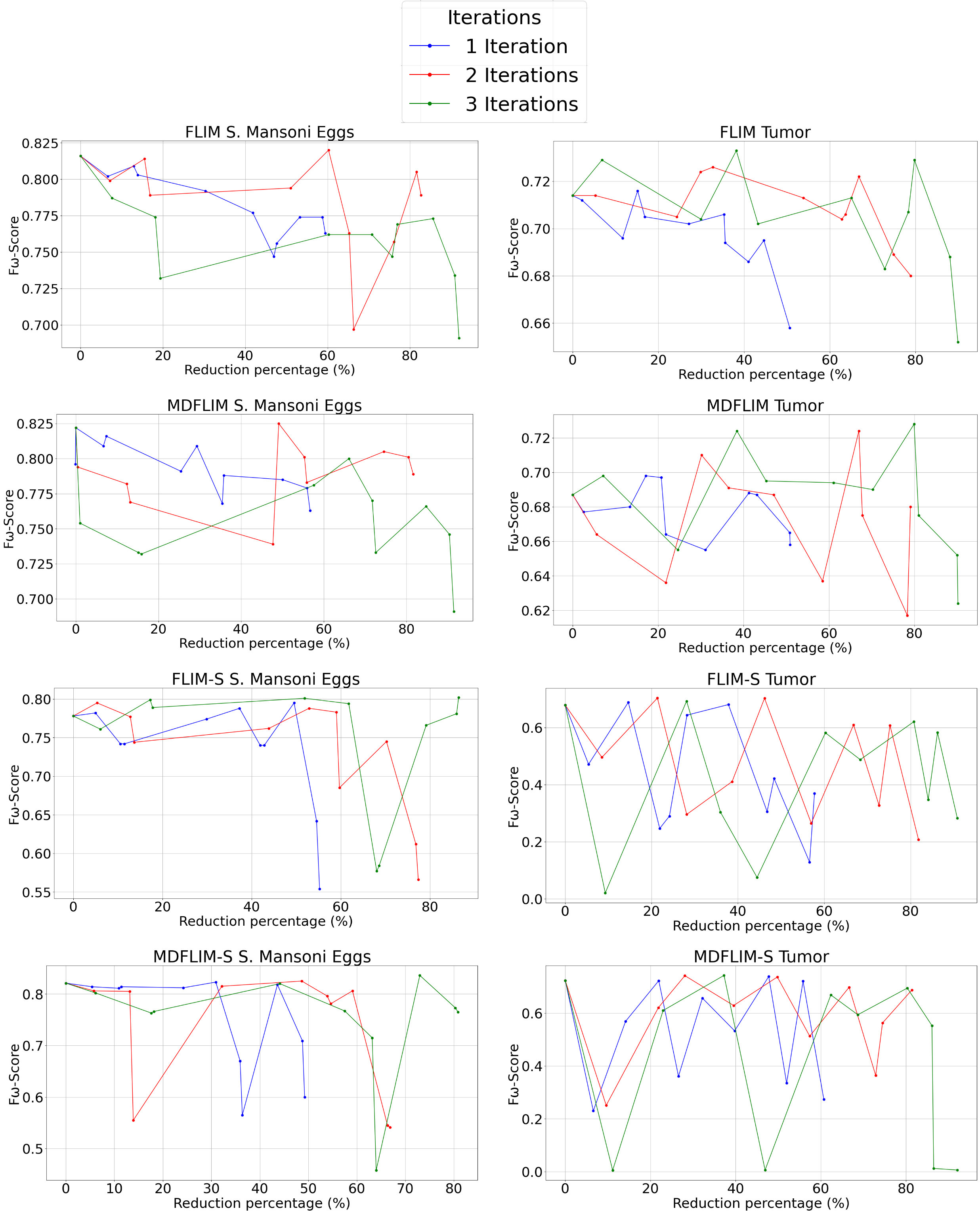} 
    \caption{Curves of $F^\omega_\beta$ over model size reduction (in number of parameters). The legend on the top discriminates to which network type each graph belongs to.}
   \label{fig:simplification-graphs}
\end{figure}

The simplification results are not deterministic with respect to parameter choices, as illustrated in Figure \ref{fig:simplification-graphs}. The left column presents all the curves for the \textit{S. mansoni} eggs dataset, while the right column shows all the curves for the Tumor dataset. There is no observable correlation between model size reduction and $F^\omega_\beta$ loss. For the \textit{S. mansoni} eggs dataset, the variance is less pronounced compared to the Tumor dataset, with most parameter choices resulting in a performance loss of approximately $0.05$ points. In contrast, for the Tumor dataset, simplification is highly sensitive to parameter choices, with the mean $F^\omega_\beta$ ranging from $0.0$ to $0.7$.

Regarding training times, Table \ref{tab:times-train} presents the average time in seconds required to train each model from scratch, as well as the cumulative time to train and perform multiple simplification iterations for each model type (the reported simplification times account for the simplification of every layer). These results represent the mean of five executions for each setup across both datasets.

\begin{table}[!h]
 \centering
    \resizebox{0.8\linewidth}{!}{%
    \begin{tabular}{l|c|c|c|c}
    \hline
    \rowcolor{gray!75}\textcolor{blue}{\textbf{\textit{S. mansoni Eggs}}} 
                &Train  &Simplify (1)   & Simplify (2)  & Simplify (3)\\ \hline 
    FLIM (GPU)  &2.26s  &6.67s          &9.96s          &12.73s\\ \hline 
    FLIM-S (GPU)&3.09s  &8.38s          &12.17s         &15.42s\\ \hline 
    FLIM (CPU)  &2.50s  &7.24s          &10.79s         &13.76s\\ \hline 
    FLIM-S (CPU)&3.60s  &9.97s          &14.71s         &18.49s\\ \hline 
    \end{tabular}
    }
    \caption{Cumulative time for training with and without simplification in GPU and CPU.}
    \label{tab:times-train}
\end{table}

The fastest training setup was for the regular FLIM, which took an average of $2.26$ seconds on the GPU, while the slowest was training separable FLIM on the CPU, adding slightly more than one second to the training time. Regarding the simplification steps, the slowest on average was simplifying separable models on the CPU, which added slightly more than six seconds to the training. Each simplification iteration required less time to execute, as there were fewer kernels to process.

For the cross-validation results, Table \ref{tab:val-flim-results} presents the mean and standard deviation for each network type across all three splits. Overall, the standard deviations are low among splits, except for the separable models on the Tumor dataset.
    
\begin{table}[!h]
 \centering
    \resizebox{0.8\linewidth}{!}{%
    \begin{tabular}{l|c|c|c|c}
    \hline
    \rowcolor{gray!75}\textcolor{blue}{\textbf{\textit{S. mansoni} eggs}} 
                    &\#Params    &FLOPs(G)   & $F^\omega_\beta$   & MAE\\ \hline 
    FLIM            &12.73(K)   &0.69   &0.789$\pm$0.022&1.139$\pm$0.162\\ \hline
    \textbf{MDFLIM} &12.73(K)   &2.08   &0.805$\pm$0.013&1.214$\pm$0.129\\ \hline
    \textbf{FLIM-S} &\textbf{2.21(K)}   &\textbf{0.11} &0.733$\pm$0.044&1.076$\pm$0.079\\ \hline
    \rowcolor{green!35}
    \textbf{MDFLIM-S}  &\textbf{2.21(K)}   &0.34 &\textbf{0.801}$\pm$0.024&\textbf{0.887}$\pm$0.116\\ \hline
    \hline 
    
    \rowcolor{gray!75}
    \textcolor{blue}{\textbf{Tumor}}
                    &\#Params    &FLOPs(G)   & $F^\omega_\beta$   & MAE\\ \hline 
    FLIM            &41.06(K)   &0.58 &\textbf{0.690}$\pm$0.014&2.534$\pm$0.783\\ \hline
    \textbf{MDFLIM} &41.06(K)   &1.74  &0.658$\pm$0.027&2.888$\pm$0.511\\ \hline
    \textbf{FLIM-S} &\textbf{5.81(K)} &\textbf{0.08} &0.229$\pm$0.317&0.958$\pm$1.120\\ \hline
    \rowcolor{green!35}
    \textbf{MDFLIM-S} &\textbf{5.81(K)} &0.238 &0.685$\pm$0.027&\textbf{2.416}$\pm$0.463\\ \hline
    \end{tabular}
    }
    \caption{Mean quantitative results for all splits in both datasets.}
    \label{tab:val-flim-results}
\end{table}

\subsubsection{Discussion and Qualitative Results}\label{sec:discuss-flim}

\begin{figure}[]
    \centering
         \includegraphics[width=0.9\textwidth]{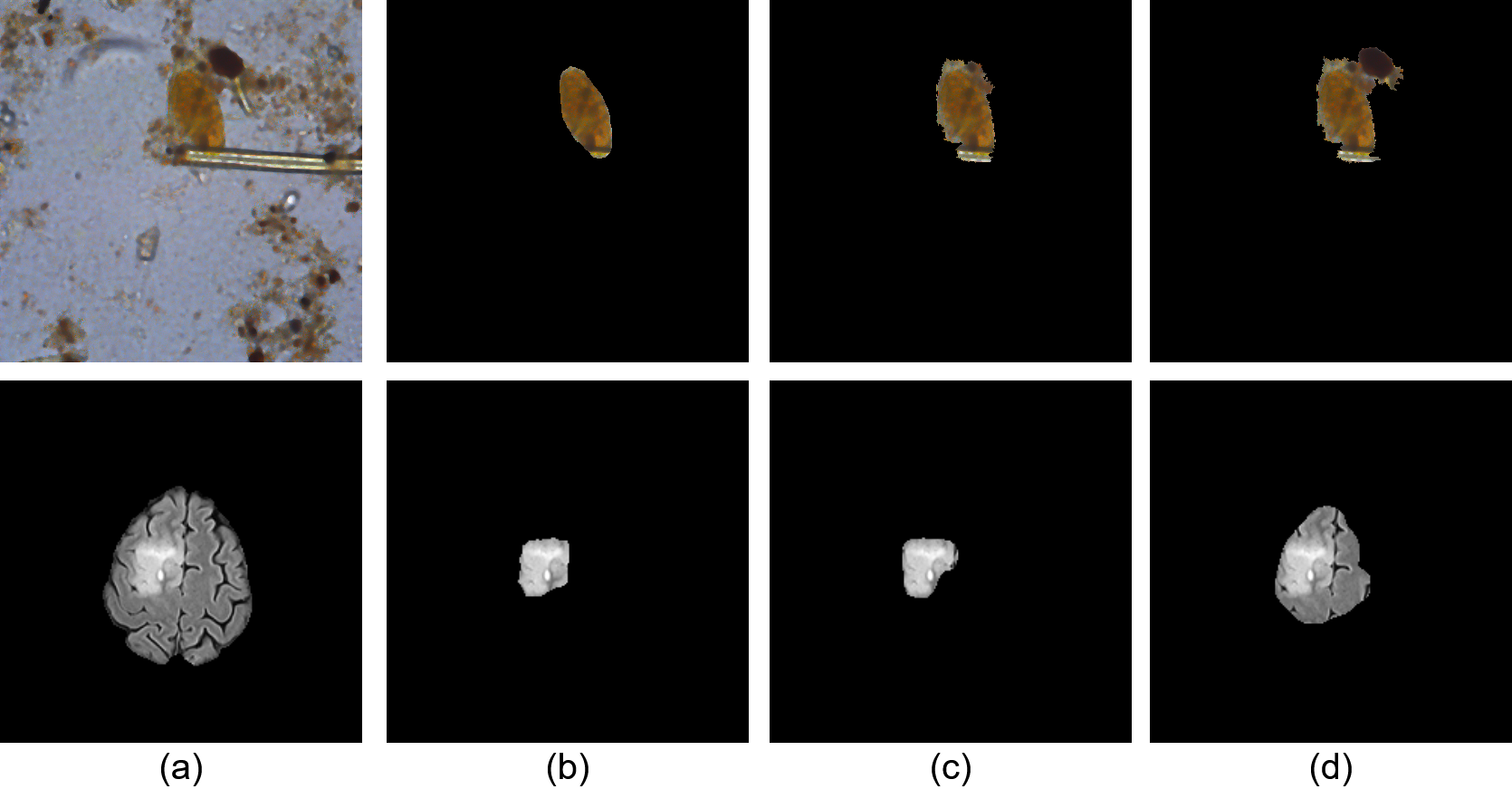} 
    \caption{(a) Original Image; (b) Ground-truth; (c-d) Results of FLIM and MDFLIM, respectively.}
   \label{fig:regular-dilated-separable}
\end{figure}

Based on the results presented in Table \ref{tab:test-flim-results}, which address the performance decrease observed when adding multi-dilation layers to regular FLIM CNNs, Figure \ref{fig:regular-dilated-separable} illustrates examples of each model's behavior on both datasets. For regular MDFLIM, other background structures are being detected as objects. A visual analysis of the output indicates there is a larger number of true positive components, but the delineation quality and non-filtered false positive hinder the saliency metrics.

\begin{figure}[]
    \centering
         \includegraphics[width=1.0\textwidth]{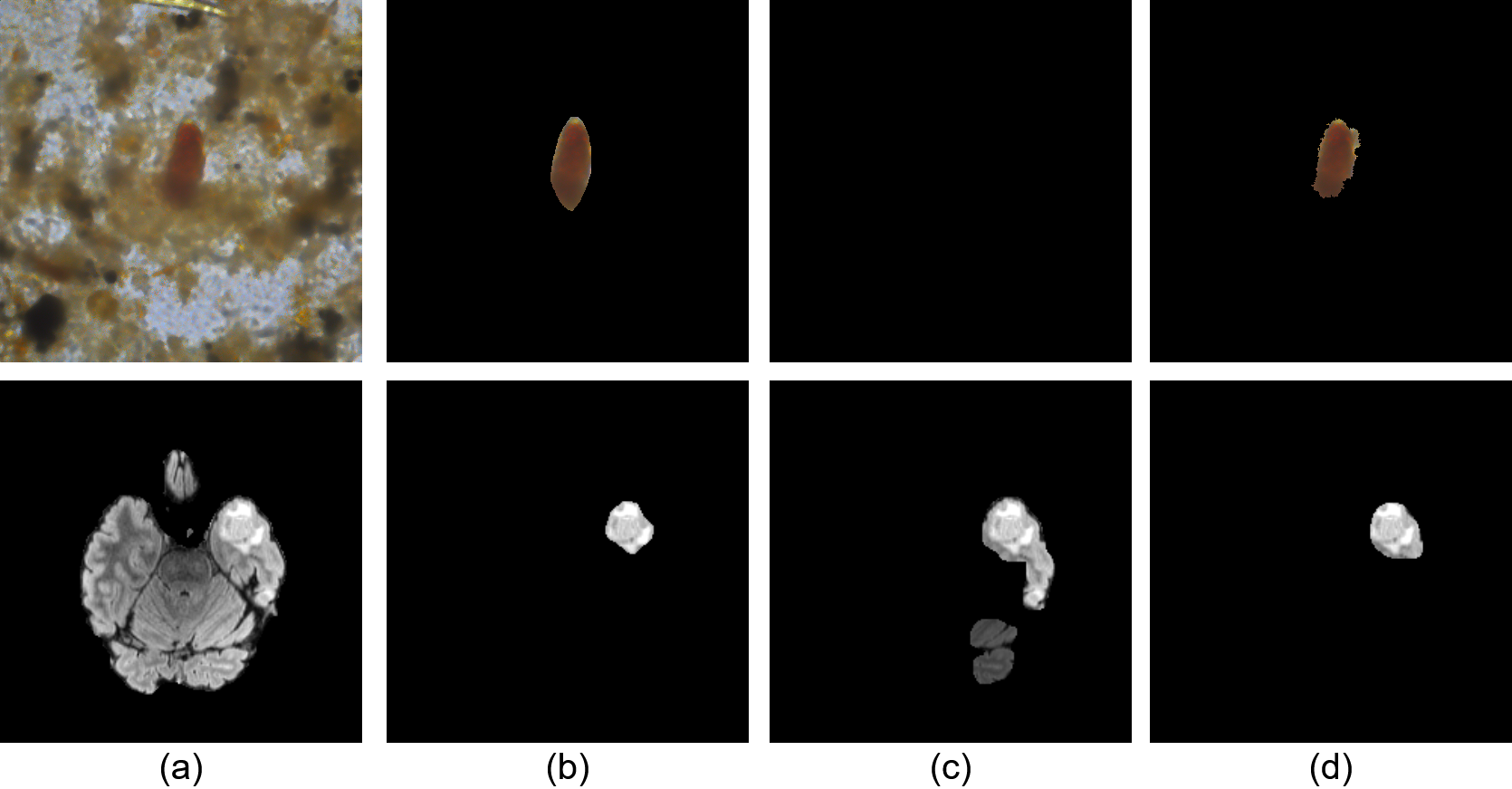} 
    \caption{(a) Original Image; (b)Ground-truth; (c-d) Results of FLIM-S and MDFLIM-S , respectively.}
   \label{fig:separable-dilated-separable}
\end{figure}

For the addition of multi-dilation layers to separable models, Figure \ref{fig:separable-dilated-separable} presents one example from each dataset to illustrate how the model improved. In the \textit{S. mansoni} eggs dataset, separable models often fail to detect the object of interest in highly cluttered images, particularly when impurities are connected to the parasite. In the Tumor dataset, most improvements stem from a reduction in false positives. In both scenarios, multi-scale features were essential to ensure proper object representation.

The differences in model size and number of operations are substantial when comparing FLIM and FLIM-S models, while the reduction in performance metrics remains mild. For MDFLIM-S, there was a significant performance gain compared to both FLIM and FLIM-S, with the number of operations and model size still smaller than those of regular FLIM. These results suggest that utilizing lightweight operations in FLIM networks is advantageous, and further exploration of alternative mechanisms to learn them should be explored.

\begin{figure}[]
    \centering
         \includegraphics[width=0.8\textwidth]{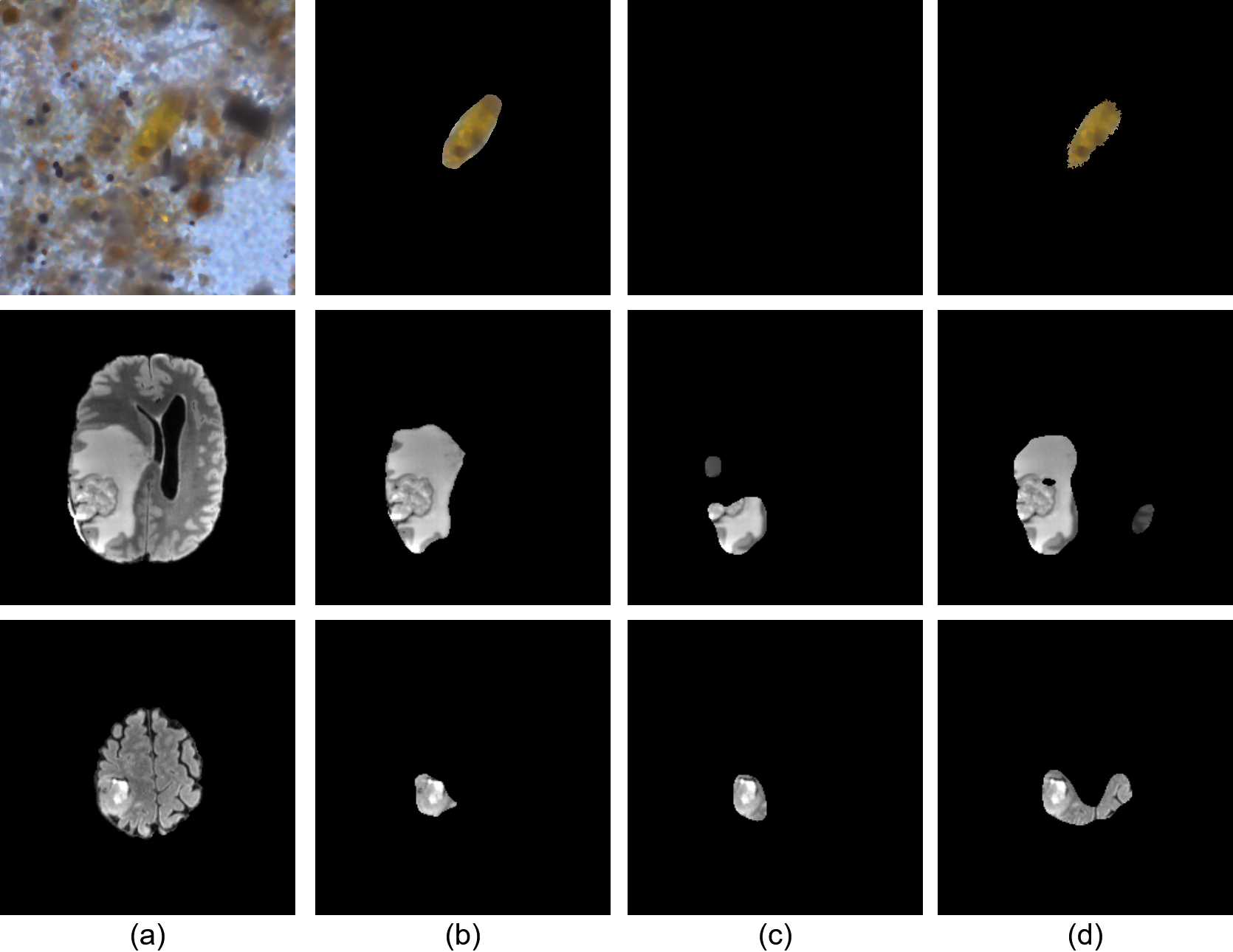} 
    \caption{(a) Original Image; (b)Ground-truth; (c-d) Results of MDFLIM-S and MDFLIM-S* models, respectively.}
   \label{fig:with-without-simp}
\end{figure}

The statistical significance test shows that most network types produce different outputs even when trained under the same regime, using the same images and base architecture. Regarding the differences between the models and their simplified versions, on the \textit{S. mansoni Eggs} dataset, most simplifications result in statistically similar performance, except for MDFLIM. In the Tumor data set, only MDFLIM did not show statistical difference. For the ones with statistical relevance, an evaluation of the saliency results shows that the simplified version occasionally detect more objects, leading to an increase in either true positives (rows one and two in Figure~\ref{fig:with-without-simp}) or false negatives (row three). Although the results are not identical, the change in performance is relatively small, while the efficiency gains are significant.

The overall small standard deviations among splits for the \textit{S. mansoni Eggs} dataset indicate that the methodology can consistently provide a suitable solution for different training image selections, provided the training set is representative. However, for a more challenging task, such as the Tumor dataset, the variation is larger, suggesting that a robust strategy for image selection remains crucial. Future work could benefit from developing an automatic or assisted image selection strategy.

\begin{figure}[]
    \centering
         \includegraphics[width=1.0\textwidth]{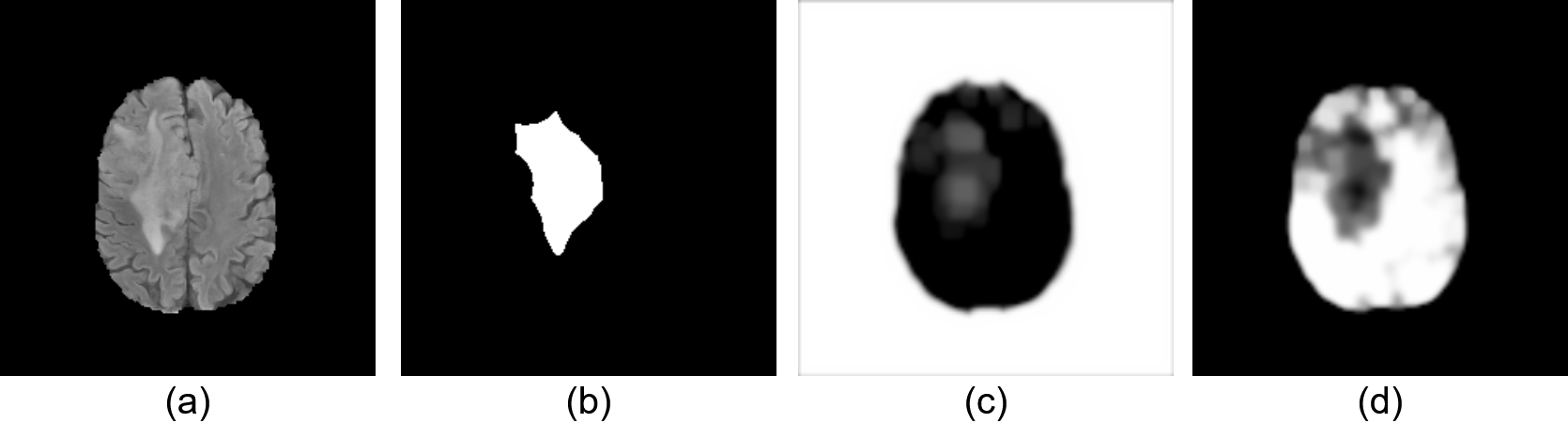} 
    \caption{(a) Original Image; (b) Ground-truth; (c-d) Results of a foreground and background kernel, respectively, for FLIM-S on a poor performing split.}
   \label{fig:val-norm-issue}
\end{figure}

Notably, FLIM-S exhibited poor average performance on the Tumor dataset, with a large standard deviation. While one split achieved results comparable to other network types, two splits failed almost completely. An evaluation of the activation maps revealed that, although good features were being extracted, they were activated along with the black background due to normalization issues (Figure \ref{fig:val-norm-issue}). Feature maps like the one shown in Figure \ref{fig:val-norm-issue}.(c) are highly detrimental to the adaptive decoder. A more complex or robust decoder could be less susceptible to these errors, potentially reducing the variation between splits.

\subsection{Comparison with state-of-the-art SOD methods}\label{subsec:comparison-sota}
In Section \ref{sec:results-sota}, we present the quantitative results for the test set across all comparing methods (heavy and lightweight) and the best flyweight model variations. Next, we present the methods performance without post-processing for the \textit{S. mansoni} eggs dataset. Finally, in Section \ref{sec:discuss-sota}, we show qualitative results, present discussions, conclusions and explanations for the results.

\subsubsection{Results}\label{sec:results-sota}
Table~\ref{tab:test-results-sota} presents two state-of-the-art (SOTA) heavyweight SOD methods alongside three SOTA LW models, as well as the best regular and separable FLIM-based models from Section~\ref{sec:results-flim}.

For the Tumor dataset, the lightweight models failed to learn a suitable representation, achieving very low $F^\omega_\beta$ scores and high MAE values, rendering them incomparable to the other models. Specifically, MSCNet exhibited extremely high MAE scores, with results approximately 20 times larger than those of the best models. In contrast, when compared to heavyweight models, the flyweight models achieved comparable and competitive performance results for both $F^\omega_\beta$ and MAE.

In terms of the number of parameters, the regular simplified FLIM model is more than 150 times smaller than the smallest CNN (SAMNet) and requires less than half the operations. Meanwhile, \textbf{MDFLIM-S}* is over 400 times smaller than SAMNet, with approximately 4 times fewer operations. When comparing the best FLIM model to heavyweight networks, the number of parameters is more than 14,000 times smaller than U²Net and 28,000 times smaller than BASNet, requiring 909 times fewer operations than BASNet and 309 times fewer than U²Net.

For the \textit{S. mansoni} eggs dataset, \textbf{MDFLIM-S}* achieved $F^\omega_\beta$ results comparable to the best-performing lightweight models, but with significantly better efficiency. The only lightweight model with a similar FLOP count (SAMNet) showed much lower performance, with an $F^\omega_\beta$ score almost 0.2 points lower and approximately three times the number of FLOPs. Even when comparing the best-performing lightweight models to the regular simplified FLIM, the $F^\omega_\beta$ difference is less pronounced than the reduction in the number of operations.

When comparing our solutions with the heavyweight models, U²Net and BASNet presented an improvement in $F^\omega_\beta$ for the \textit{S. mansoni} eggs dataset, and MAE improvement on the Tumor one. However, the best presented flyweight model have substantially smaller MAE than any other method for the \textit{S. mansoni} eggs dataset, and on-par $F^\omega_\beta$ in the brain tumor data set. 

\begin{table}[!h]
 \centering
    \resizebox{0.6\linewidth}{!}{%
    \begin{tabular}{l|c|c|c|c}
    \hline
    \rowcolor{gray!75}\textcolor{blue}{\textbf{\textit{S. mansoni Eggs}}} 
                    &\#Params    &FLOPs(G)   & $F^\omega_\beta$   & MAE \\ \hline 
                    
    \rowcolor{gray!35}
    BASNet          &87.06(M)   &127.3      &0.850 & 1.000\\ \hline
    \rowcolor{gray!35}
    U$^2$Net        &\textbf{44.3(M)}    &\textbf{58.80}      &\textbf{0.867}&\textbf{0.773}\\ \hline
    MEANet          &3.27(M)    &5.87    &0.832&1.645\\ \hline
    MSCNet          &3.26(M)    &9.62    &0.832&1.335\\ \hline
    SAMNet          &\textbf{1.33(M)}    &\textbf{0.5} &0.634&2.681\\ \hline
    FLIM*           &7.06(K)   &0.39&0.820&0.822\\ \hline
    \rowcolor{green!35}
    \textbf{MDFLIM-S}*&1.15(K) &0.19 &\textbf{0.837}&\textbf{0.484}\\ \hline
    \hline 
    
    \rowcolor{gray!75}
    \textcolor{blue}{\textbf{Tumor}}
                    &\#Params    &FLOPs(G)   & $F^\omega_\beta$   & MAE\\ \hline 
    \rowcolor{gray!35}
    BASNet          &87.06(M)   &127.3      &0.724 & \textbf{1.893}\\ \hline
    \rowcolor{gray!35}
    U$^2$Net        &\textbf{44.3(M)}    &\textbf{58.80}      &\textbf{0.739}&2.065\\ \hline
    MEANet          &3.27(M)    &5.87       &0.128&6.218\\ \hline
    MSCNet          &3.26(M)    &9.62       &0.236&37.368\\ \hline
    SAMNet          &\textbf{1.33(M)}    &\textbf{0.5} &0.141&9.115\\ \hline
    FLIM*           &8.12(K)   &0.21 &0.723&\textbf{2.447}\\ \hline
    \rowcolor{green!35}
    \textbf{MDFLIM-S}*&3.13(K) &0.14 &\textbf{0.739}&2.469\\ \hline
    \end{tabular}
    }
    \caption{Quantitative results. The best results for lightweight, heavy, and proposed methods are in bold. Lightweight rows are in white, and heavy in gray. The proposed methods are in bold.}
    \label{tab:test-results-sota}
\end{table}

For the \textit{S. mansoni Eggs} dataset, Table~\ref{tab:test-results-without-filter} presents the results of each method without any post-processing. While all methods benefit significantly from the post-processing step, the FLIM-based models exhibit the greatest performance improvement, especially for MAE.

\begin{table}[!h]
 \centering
    \resizebox{0.7\linewidth}{!}{%
    \begin{tabular}{l|c|c}
    \hline
    \rowcolor{gray!75}\textcolor{blue}{\textbf{\textit{S. mansoni Eggs}}} 
                    & $F^\omega_\beta$   & MAE\\ \hline 
                    
    \rowcolor{gray!35}
    BASNet          &0.365 (-57.1\%) & 2.304 (130.4\%)\\ \hline
    \rowcolor{gray!35}
    U$^2$Net        &0.385 (-55.6\%) & 1.377 (78.13\%)\\ \hline
    MEANet          &0.380 (-54.3\%) &2.080 (26.4\%) \\ \hline
    MSCNet          &0.352 (-57.7\%) &1.974 (47.86\%) \\ \hline
    SAMNet          &0.302 (-52.36\%) &4.272 (159.34\%) \\ \hline
    \textbf{FLIM}*  &0.350 ( -57.32\%) & 2.954 (259.36\%)\\ \hline
    \textbf{MDFLIM-S}*&0.364 ( -56.51\%) & 2.080 (329.75\%)\\ \hline
    \end{tabular}
    }
    \caption{Non-filtered results for the \textit{S. mansoni Eggs} dataset and the difference of measure value from the post-processed result.}
    \label{tab:test-results-without-filter}
\end{table}

\subsubsection{Discussion and qualitative results}\label{sec:discuss-sota}
\begin{figure}[]
        \centering
         \includegraphics[width=1.0\textwidth]{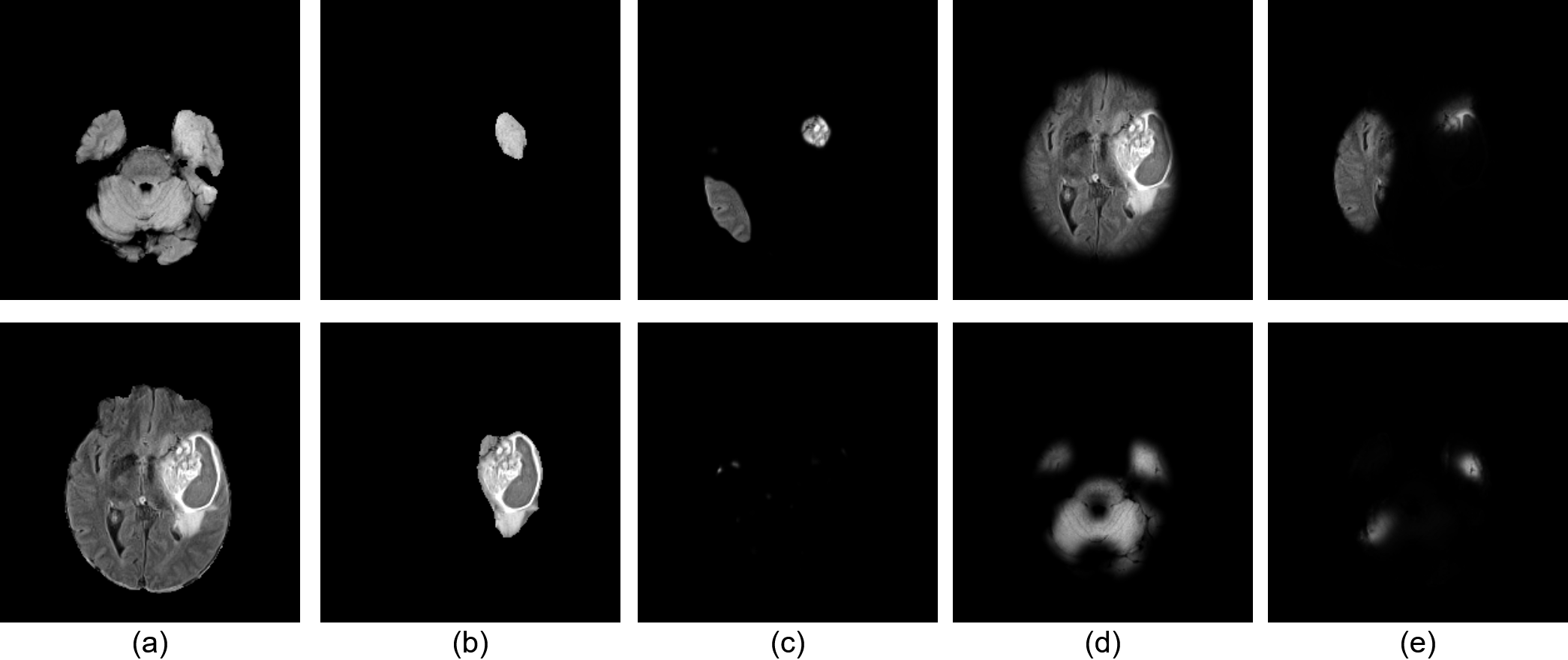} 
        \caption{Example of lightweight method results on the Tumor dataset. (a) Original Image; (b)Ground-truth; (c) MEANet; (d) MSCNet; (e) SAMNet.}
       \label{fig:lightweight-brain}
    \end{figure}

    Building on the underperformance of lightweight models on the Tumor dataset, Figure \ref{fig:lightweight-brain} illustrates the results of each method. The extremely high MAE score for MSCNet is attributed to outcomes like the example shown in Figure \ref{fig:lightweight-brain}(d). The underperformance of other methods is often linked to their inability to detect the correct object.
    
    Because lightweight models have considerably fewer parameters (and thus features) than heavyweight ones, their features must be far more specialized to the training dataset to achieve good results. When shifting between such different domains (e.g., from colored natural images to grayscale MRIs), the less generic features of the lightweight models could not adapt effectively during fine-tuning with such a limited number of images, resulting in poor performance. FLIM-based methods, by contrast, learn the model directly for the task instead of relying on fine-tuning, enabling the creation of very small and specialized models without sacrificing performance.

\begin{figure}[]
        \centering
         \includegraphics[width=1.0\textwidth]{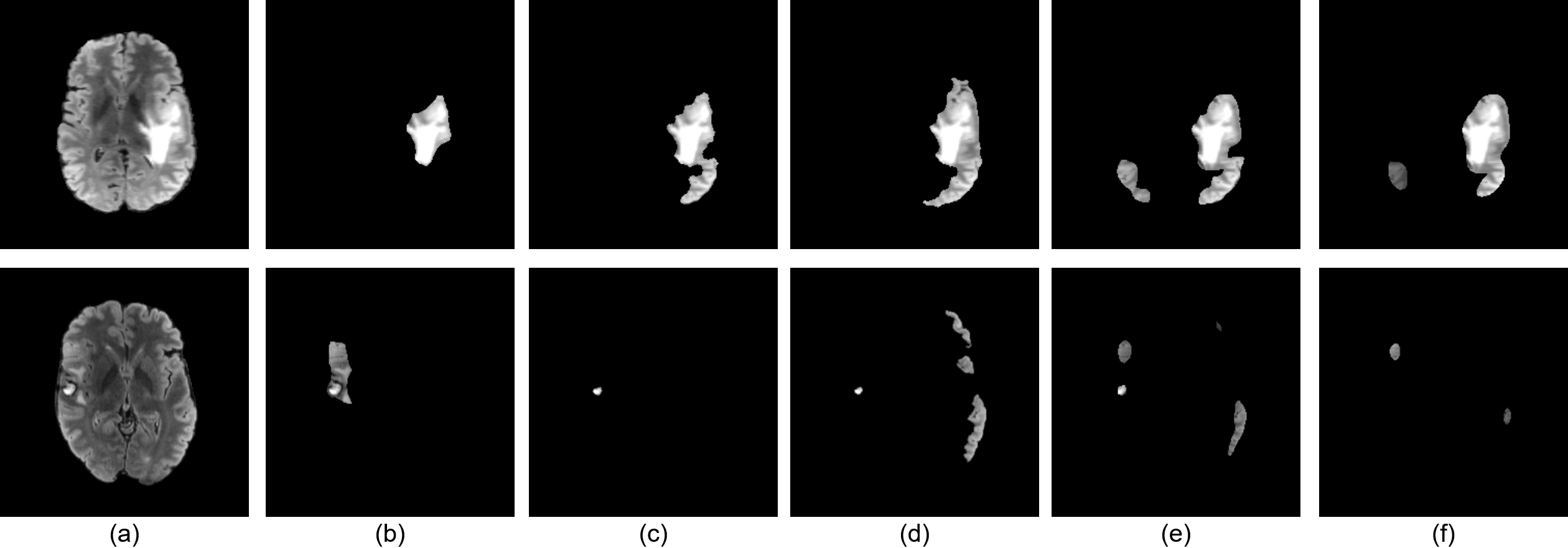} 
        \caption{Examples of similar performance from both heavyweight and flyweight methods for the Tumor dataset. (a) Original Image; (b)Ground-truth; (c) Basnet; (d) U²Net; (e) FLIM*; (f) MDFLIM-S*.}
       \label{fig:brain-heavy-all-bad}
    \end{figure}

    Compared to heavy-weight methods, FLIM achieves similar performance. Figure~\ref{fig:brain-heavy-all-bad} illustrates examples of misrepresented salient objects. In the first row, all models partially detected the same brain structures (tumor and false positives). In the second row, the correct salient object was almost entirely missed by the heavyweight models and MD-FLIM-S, while FLIM detected the tumor but also highlighted non-tumor connected regions. Despite the similarity in results, the FLIM-based methods produce binary saliency maps with less pronounced saliency for non-salient objects.

\begin{figure}[]
        \centering
         \includegraphics[width=1.0\textwidth]{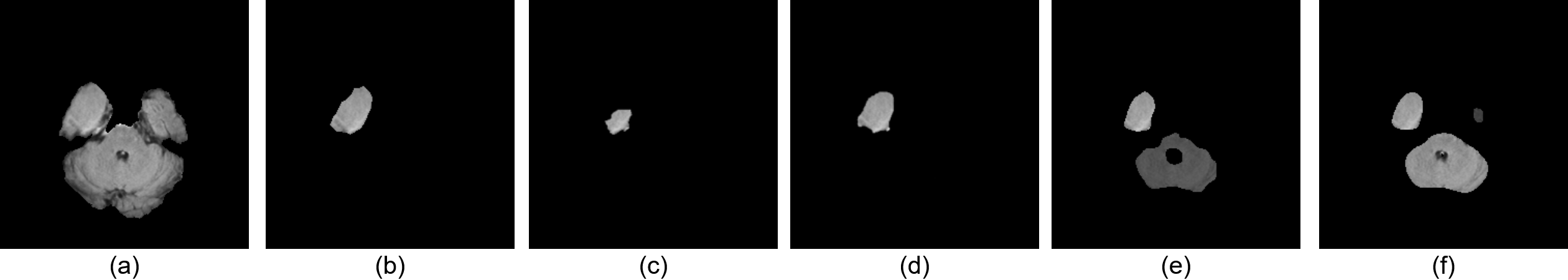} 
        \caption{Examples of poor performance from flyweight methods compared to good heavyweight models for the Tumor dataset. (a) Original Image; (b)Ground-truth; (c) Basnet; (d) U²Net; (e) FLIM*; (f) MDFLIM-S*.}
       \label{fig:brain-heavy-flim-bad}
    \end{figure}

    On images with a very subtle contrast between tumors and healthy tissues, deep learning methods often outperform FLIM-based approaches, as illustrated in Figure \ref{fig:brain-heavy-flim-bad}. In such cases, leveraging deeper features appears to be essential.

\begin{figure}[]
        \centering
         \includegraphics[width=1.0\textwidth]{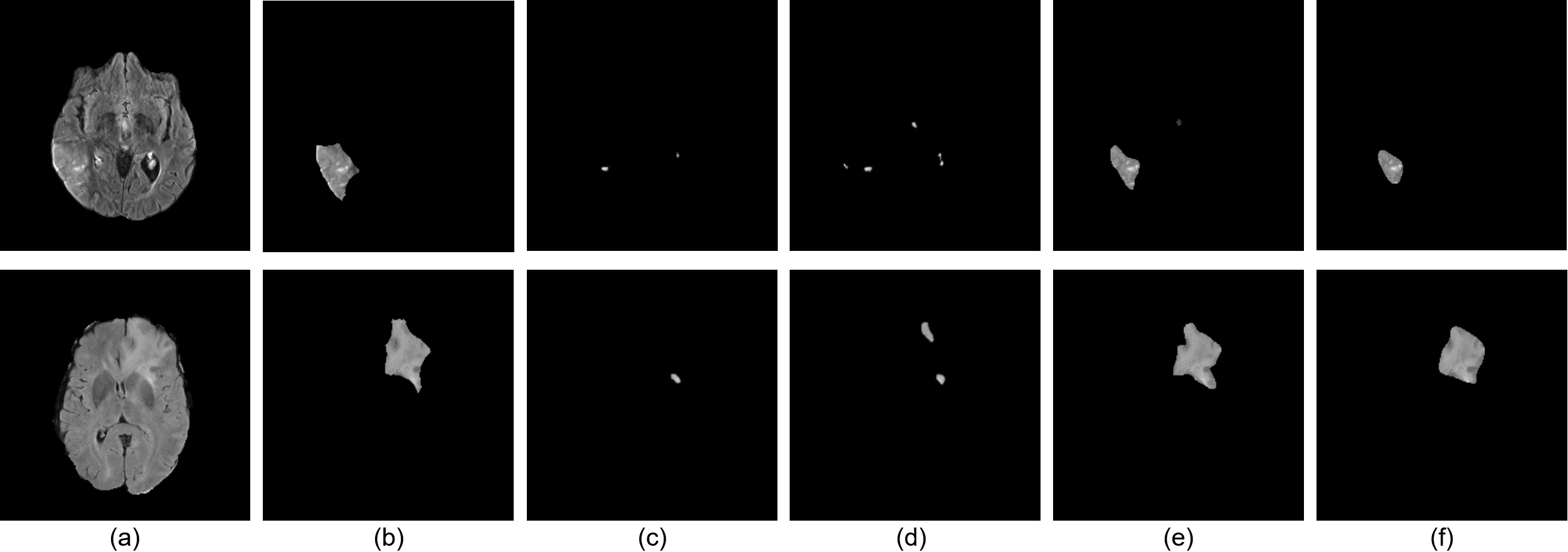} 
        \caption{Examples of satisfactory performance from flyweight methods on images with poor performance from heavyweight models for the Tumor dataset. (a) Original Image; (b)Ground-truth; (c) Basnet; (d) U²Net; (e) FLIM*; (f) MDFLIM-S*.}
       \label{fig:brain-heavy-flim-good}
    \end{figure}

    However, deep features are sometimes unsuitable for images with a clear visual distinction between tumors and healthy tissues, as shown in Figure~\ref{fig:brain-heavy-flim-good}. In such cases, since FLIM-based methods rely on features learned directly for the target task, their results are often more reliable for visually distinct objects.

\begin{figure}[]
        \centering
         \includegraphics[width=1.0\textwidth]{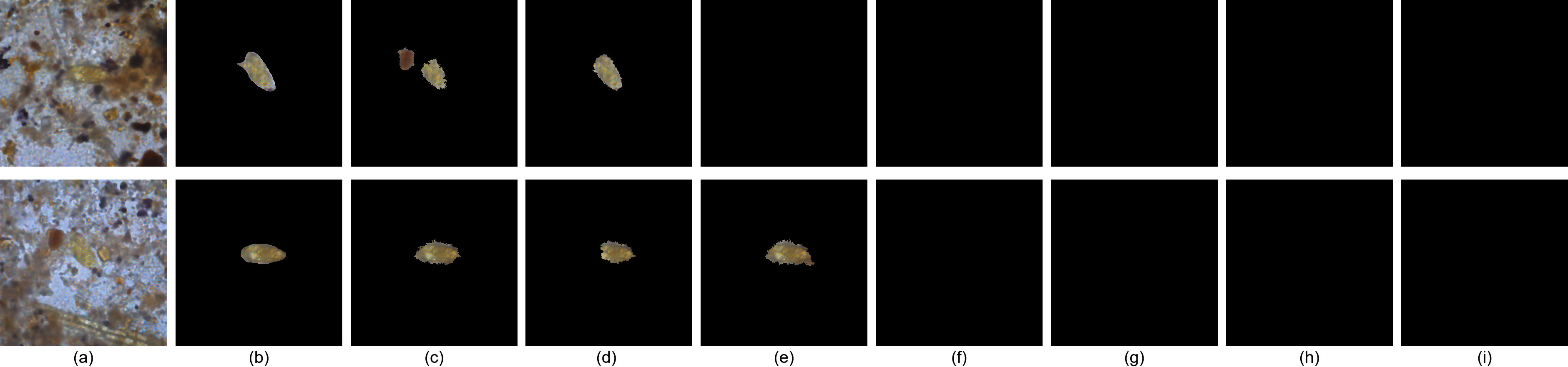} 
        \caption{Examples of bad performance from flyweight and Lightweight methods for the \textit{S. mansoni Eggs} dataset. (a) Original Image; (b)Ground-truth; (c) Basnet; (d) U²Net; (e) MEANet; (f) MSCNet; (g) SAMNet; (h) FLIM*; (i) MD-FLIM-S*.}
       \label{fig:parasites-flim-bad}
    \end{figure}

    Regarding the \textit{S. mansoni} eggs dataset and the superior performance of heavyweight models, Figure~\ref{fig:parasites-flim-bad} illustrates examples where lightweight and flyweight methods perform poorly. In both images (first and second rows), the object of interest (parasite) has weakly defined borders, shares similarities with other objects (impurities), and is connected to or obscured by impurities. For these images, BASNet and U²Net seem to benefit from their very deep features, whereas the lightweight and flyweight models fail to detect any objects (except for MSCNet in the first row).

\begin{figure}[]
        \centering
         \includegraphics[width=1.0\textwidth]{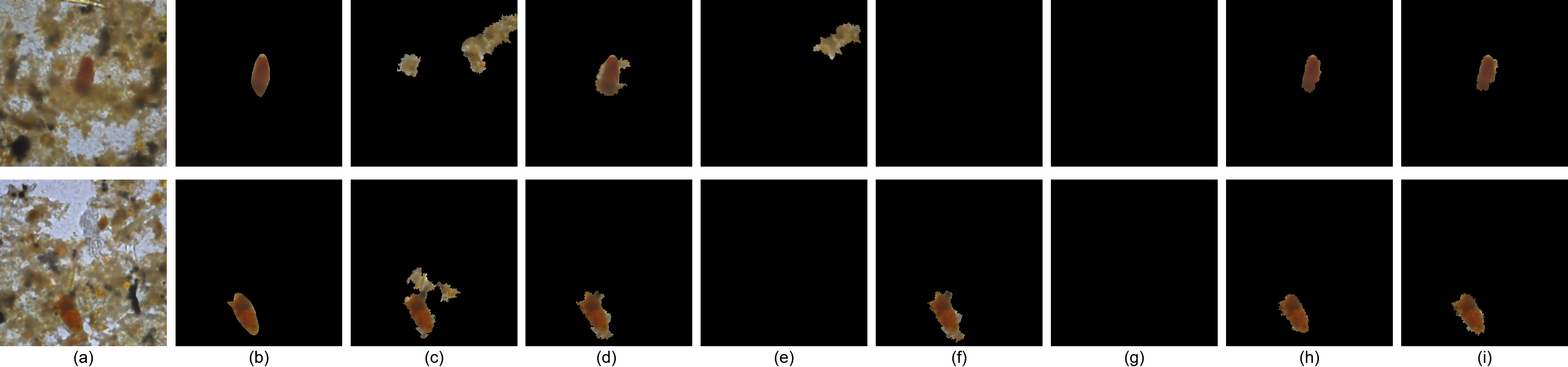} 
        \caption{Examples of good performance from flyweight methods and poor performance from some lightweight models for the \textit{S. mansoni Eggs} dataset. (a) Original Image; (b)Ground-truth; (c) Basnet; (d) U²Net; (e) MEANet; (f) MSCNet; (g) SAMNet; (h) FLIM*; (i) MD-FLIM-S*.}
       \label{fig:parasites-flim-good}
    \end{figure}

    For FLIM-based methods, the adaptive decoder uses a heuristic based on the mean saliency to distinguish between background and foreground filters. However, the presence of numerous or large impurities that share characteristics with the parasite can be highly detrimental, causing foreground activations to be misclassified as background. Conversely, on heavily cluttered images where the parasite eggs exhibit distinct characteristics from most impurities, the flyweight models can produce accurate saliency results, as shown in Figure~\ref{fig:parasites-flim-good}.

\begin{figure}[]
        \centering
         \includegraphics[width=1.0\textwidth]{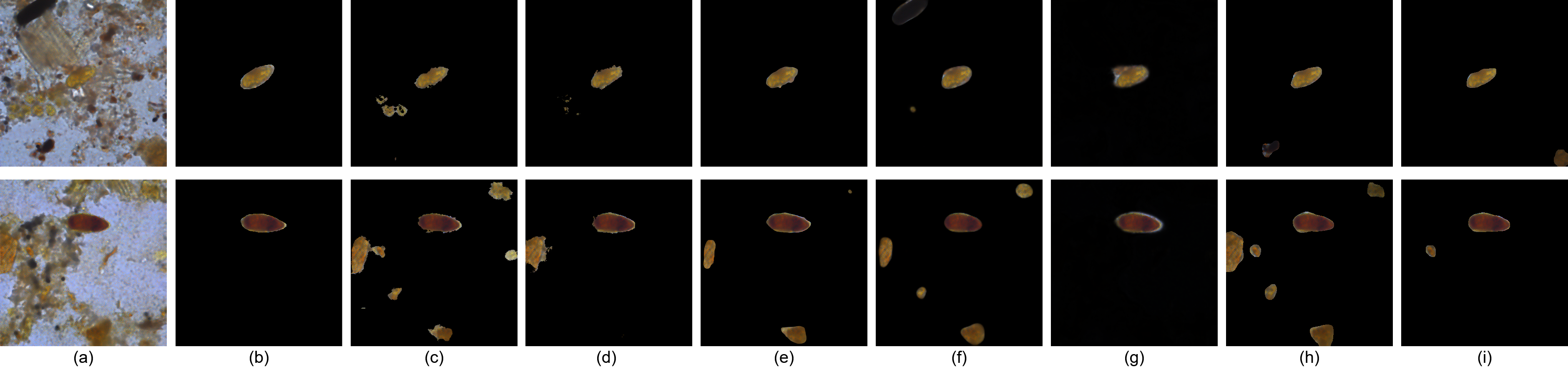} 
        \caption{Results of each method without post-processing steps for the \textit{S. mansoni Eggs} dataset. (a) Original Image; (b)Ground-truth; (c) Basnet; (d) U²Net; (e) MEANet; (f) MSCNet; (g) SAMNet; (h) FLIM*; (i) MD-FLIM-S*.}
       \label{fig:parasites-non-filtered}
    \end{figure}

    In terms of post-processing usage, Figure~\ref{fig:parasites-non-filtered} presents examples of non-filtered results for all methods. All methods benefit from the area filter, which reduces the number of false positives, but the impact is more pronounced for the flyweight methods (last two columns), indicating that more false positives in FLIM's results are very small or very large components, which are easily filtered out. Additionally, note that non-flyweight methods typically produce almost binary results with well-defined borders, whereas FLIM-based methods often under-represent object sizes. Since backpropagation-trained models aim to approximate the binary ground truth using losses that favor boundary adherence, their outputs closely resemble segmentation maps. Consequently, FLIM-based methods gain significant advantages from segmentation post-processing, which brings the saliency results closer to the binary ground truth.

\section{Conclusion}\label{sec:conclusion}
    We presented a methodology for learning flyweight models using FLIM, incorporating techniques for network simplification and the learning of lightweight operations (dilated separable convolutions) within the FLIM framework. FLIM-based models are simplified layer by layer by removing redundant kernels. Separable FLIM layers are trained by extending FLIM's traditional learning strategy to include depthwise and pointwise factorization using kernel bank statistics. Additionally, we introduced support for multi-dilation layers in both separable and regular FLIM CNNs. The proposed CNNs eliminate the need for backpropagation by leveraging unsupervised adaptive decoders, which can now explore information at multiple scales through the use of multi-dilation layers. Our results showed significant improvements in performance and efficiency across four metrics compared to baseline FLIM networks. The best flyweight models achieved superior performance and efficiency when compared to lightweight SOD models and achieved competitive performance with greatly improved efficiency compared to heavyweight state-of-the-art (SOTA) SOD models.

    For future work, other aspects of FLIM could be improved, such as developing an image selection strategy to improve robustness to changes in the training set. Additionally, exploring other lightweight operations, employing more robust decoders, and evaluating the approach on diverse datasets and domains, such as remote sensing images, would be valuable directions for further research.

\section*{Acknowledgements}\label{sec:acks}
´This work was financially supported by the Conselho Nacional de Desenvolvimento Cient{\'i}fico e Tecnol{\'o}gico -- CNPq -- (Universal 407242/2021-0, International 442950/2023-3, PQ 306573/2022-9, PQ 304711/2023-3, 310075/2019-0), the Funda\c{c}{\~a}o de Amparo a Pesquisa do Estado de Minas Gerais -- FAPEMIG -- (PPM-00006-18, APQ-01079-23 and APQ-05058-23), the Funda\c{c}{\~a}o de Amparo a Pesquisa do Estado de S{\~a}o Paulo -- FAPESP -- (CPA-FAPESP 2023/14427-8) and the Coordena{\c c}{\~a}o de Aperfei{\c c}oamento de Pessoal de N{\'i}vel Superior -- CAPES -- Finance code 001 (COFECUB 88887.191730/2018-00 and STIC-AMSUD 88887.878869/2023-00).

\printcredits

\clearpage

\bibliographystyle{cas-model2-names}

\bibliography{main}

\begin{thebibliography}{31}
\expandafter\ifx\csname natexlab\endcsname\relax\def\natexlab#1{#1}\fi
\providecommand{\url}[1]{\texttt{#1}}
\providecommand{\href}[2]{#2}
\providecommand{\path}[1]{#1}
\providecommand{\DOIprefix}{doi:}
\providecommand{\ArXivprefix}{arXiv:}
\providecommand{\URLprefix}{URL: }
\providecommand{\Pubmedprefix}{pmid:}
\providecommand{\doi}[1]{\href{http://dx.doi.org/#1}{\path{#1}}}
\providecommand{\Pubmed}[1]{\href{pmid:#1}{\path{#1}}}
\providecommand{\bibinfo}[2]{#2}
\ifx\xfnm\relax \def\xfnm[#1]{\unskip,\space#1}\fi
\bibitem[{Al-Azawi(2021)}]{al2021saliency}
\bibinfo{author}{Al-Azawi, M.A.}, \bibinfo{year}{2021}.
\newblock \bibinfo{title}{Saliency-based image retrieval as a refinement to content-based image retrieval}.
\newblock \bibinfo{journal}{ELCVIA: Electronic Letters on Computer Vision and Image Analysis} \bibinfo{volume}{20}, \bibinfo{pages}{0001--15}.
\bibitem[{Borji et~al.(2019)Borji, Cheng, Hou, Jiang and Li}]{borji2019salient}
\bibinfo{author}{Borji, A.}, \bibinfo{author}{Cheng, M.M.}, \bibinfo{author}{Hou, Q.}, \bibinfo{author}{Jiang, H.}, \bibinfo{author}{Li, J.}, \bibinfo{year}{2019}.
\newblock \bibinfo{title}{Salient object detection: A survey}.
\newblock \bibinfo{journal}{Computational visual media} \bibinfo{volume}{5}, \bibinfo{pages}{117--150}.
\bibitem[{Bragantini et~al.(2018)Bragantini, Martins, Castelo-Fernandez and Falc{\~a}o}]{bragantini2018graph}
\bibinfo{author}{Bragantini, J.}, \bibinfo{author}{Martins, S.B.}, \bibinfo{author}{Castelo-Fernandez, C.}, \bibinfo{author}{Falc{\~a}o, A.X.}, \bibinfo{year}{2018}.
\newblock \bibinfo{title}{Graph-based image segmentation using dynamic trees}, in: \bibinfo{booktitle}{Iberoamerican Congress on Pattern Recognition}, \bibinfo{organization}{Springer}. pp. \bibinfo{pages}{470--478}.
\bibitem[{Cerqueira et~al.(2023)Cerqueira, Sprenger, Teixeira and Falc{\~a}o}]{cerqueira2023building}
\bibinfo{author}{Cerqueira, M.A.}, \bibinfo{author}{Sprenger, F.}, \bibinfo{author}{Teixeira, B.C.}, \bibinfo{author}{Falc{\~a}o, A.X.}, \bibinfo{year}{2023}.
\newblock \bibinfo{title}{Building brain tumor segmentation networks with user-assisted filter estimation and selection}, in: \bibinfo{booktitle}{18th International Symposium on Medical Information Processing and Analysis}, \bibinfo{organization}{SPIE}. pp. \bibinfo{pages}{202--211}.
\bibitem[{Cerqueira et~al.(2024)Cerqueira, Sprenger, Teixeira, Guimarães and Falcão}]{cerqueira2024selection}
\bibinfo{author}{Cerqueira, M.A.}, \bibinfo{author}{Sprenger, F.}, \bibinfo{author}{Teixeira, B.C.A.}, \bibinfo{author}{Guimarães, S.J.F.}, \bibinfo{author}{Falcão, A.X.}, \bibinfo{year}{2024}.
\newblock \bibinfo{title}{Interactive ground-truth-free image selection for flim segmentation encoders}, in: \bibinfo{booktitle}{2024 37th SIBGRAPI Conference on Graphics, Patterns and Images (SIBGRAPI)}, pp. \bibinfo{pages}{1--6}.
\newblock \DOIprefix\doi{10.1109/SIBGRAPI62404.2024.10716300}.
\bibitem[{De~Souza and Falc{\~a}o(2020)}]{de2020learning}
\bibinfo{author}{De~Souza, I.E.}, \bibinfo{author}{Falc{\~a}o, A.X.}, \bibinfo{year}{2020}.
\newblock \bibinfo{title}{Learning cnn filters from user-drawn image markers for coconut-tree image classification}.
\newblock \bibinfo{journal}{IEEE Geoscience and Remote Sensing Letters} .
\bibitem[{Gao et~al.(2020)Gao, Tan, Cheng, Lu, Chen and Yan}]{gao2020highly}
\bibinfo{author}{Gao, S.H.}, \bibinfo{author}{Tan, Y.Q.}, \bibinfo{author}{Cheng, M.M.}, \bibinfo{author}{Lu, C.}, \bibinfo{author}{Chen, Y.}, \bibinfo{author}{Yan, S.}, \bibinfo{year}{2020}.
\newblock \bibinfo{title}{Highly efficient salient object detection with 100k parameters}, in: \bibinfo{booktitle}{European Conference on Computer Vision}, \bibinfo{organization}{Springer}. pp. \bibinfo{pages}{702--721}.
\bibitem[{Joao et~al.(2023)Joao, Santos, Guimaraes, Gomes, Kijak and Falcao}]{joao2023flyweight}
\bibinfo{author}{Joao, L.d.M.}, \bibinfo{author}{Santos, B.M.d.}, \bibinfo{author}{Guimaraes, S.J.F.}, \bibinfo{author}{Gomes, J.F.}, \bibinfo{author}{Kijak, E.}, \bibinfo{author}{Falcao, A.X.}, \bibinfo{year}{2023}.
\newblock \bibinfo{title}{A flyweight cnn with adaptive decoder for schistosoma mansoni egg detection}.
\newblock \bibinfo{journal}{preprint arXiv:2306.14840} .
\bibitem[{Joao et~al.(2024)Joao, Cerqueira, Benato and Falcão}]{JoaoCBF24}
\bibinfo{author}{Joao, L.M.}, \bibinfo{author}{Cerqueira, M.A.}, \bibinfo{author}{Benato, B.C.}, \bibinfo{author}{Falcão, A.X.}, \bibinfo{year}{2024}.
\newblock \bibinfo{title}{Understanding marker-based normalization for flim networks}, in: \bibinfo{booktitle}{19th International Joint Conference on Computer Vision, Imaging and Computer Graphics Theory and Applications, VISIGRAPP 2024}, pp. \bibinfo{pages}{612--623}.
\newblock \DOIprefix\doi{10.5220/0012385900003660}.
\bibitem[{Liang and Luo(2023)}]{liang2023meanet}
\bibinfo{author}{Liang, B.}, \bibinfo{author}{Luo, H.}, \bibinfo{year}{2023}.
\newblock \bibinfo{title}{Meanet: An effective and lightweight solution for salient object detection in optical remote sensing images}.
\newblock \bibinfo{journal}{Expert Systems with Applications} , \bibinfo{pages}{121778}.
\bibitem[{Lin et~al.(2022)Lin, Sun, Liu, Bian, Cen and Zhou}]{lin2022lightweight}
\bibinfo{author}{Lin, Y.}, \bibinfo{author}{Sun, H.}, \bibinfo{author}{Liu, N.}, \bibinfo{author}{Bian, Y.}, \bibinfo{author}{Cen, J.}, \bibinfo{author}{Zhou, H.}, \bibinfo{year}{2022}.
\newblock \bibinfo{title}{A lightweight multi-scale context network for salient object detection in optical remote sensing images}, in: \bibinfo{booktitle}{2022 26th international conference on pattern recognition (ICPR)}, \bibinfo{organization}{IEEE}. pp. \bibinfo{pages}{238--244}.
\bibitem[{Liu et~al.(2019)Liu, Hou, Cheng, Feng and Jiang}]{liu2019simple}
\bibinfo{author}{Liu, J.J.}, \bibinfo{author}{Hou, Q.}, \bibinfo{author}{Cheng, M.M.}, \bibinfo{author}{Feng, J.}, \bibinfo{author}{Jiang, J.}, \bibinfo{year}{2019}.
\newblock \bibinfo{title}{A simple pooling-based design for real-time salient object detection}, in: \bibinfo{booktitle}{IEEE/CVF conference on computer vision and pattern recognition}, pp. \bibinfo{pages}{3917--3926}.
\bibitem[{Liu et~al.(2020)Liu, Gu, Zhang, Wang and Cheng}]{liu2020lightweight}
\bibinfo{author}{Liu, Y.}, \bibinfo{author}{Gu, Y.C.}, \bibinfo{author}{Zhang, X.Y.}, \bibinfo{author}{Wang, W.}, \bibinfo{author}{Cheng, M.M.}, \bibinfo{year}{2020}.
\newblock \bibinfo{title}{Lightweight salient object detection via hierarchical visual perception learning}.
\newblock \bibinfo{journal}{IEEE transactions on cybernetics} \bibinfo{volume}{51}, \bibinfo{pages}{4439--4449}.
\bibitem[{Liu et~al.(2021)Liu, Zhang, Bian, Zhang and Cheng}]{liu2021samnet}
\bibinfo{author}{Liu, Y.}, \bibinfo{author}{Zhang, X.Y.}, \bibinfo{author}{Bian, J.W.}, \bibinfo{author}{Zhang, L.}, \bibinfo{author}{Cheng, M.M.}, \bibinfo{year}{2021}.
\newblock \bibinfo{title}{Samnet: Stereoscopically attentive multi-scale network for lightweight salient object detection}.
\newblock \bibinfo{journal}{IEEE Transactions on Image Processing} \bibinfo{volume}{30}, \bibinfo{pages}{3804--3814}.
\bibitem[{Margolin et~al.(2014)Margolin, Zelnik-Manor and Tal}]{margolin2014evaluate}
\bibinfo{author}{Margolin, R.}, \bibinfo{author}{Zelnik-Manor, L.}, \bibinfo{author}{Tal, A.}, \bibinfo{year}{2014}.
\newblock \bibinfo{title}{How to evaluate foreground maps?}, in: \bibinfo{booktitle}{IEEE conference on computer vision and pattern recognition}, pp. \bibinfo{pages}{248--255}.
\bibitem[{Qin et~al.(2020)Qin, Zhang, Huang, Dehghan, Zaiane and Jagersand}]{qin2020u2}
\bibinfo{author}{Qin, X.}, \bibinfo{author}{Zhang, Z.}, \bibinfo{author}{Huang, C.}, \bibinfo{author}{Dehghan, M.}, \bibinfo{author}{Zaiane, O.R.}, \bibinfo{author}{Jagersand, M.}, \bibinfo{year}{2020}.
\newblock \bibinfo{title}{U2-net: Going deeper with nested u-structure for salient object detection}.
\newblock \bibinfo{journal}{Pattern recognition} \bibinfo{volume}{106}, \bibinfo{pages}{107404}.
\bibitem[{Qin et~al.(2019)Qin, Zhang, Huang, Gao, Dehghan and Jagersand}]{qin2019basnet}
\bibinfo{author}{Qin, X.}, \bibinfo{author}{Zhang, Z.}, \bibinfo{author}{Huang, C.}, \bibinfo{author}{Gao, C.}, \bibinfo{author}{Dehghan, M.}, \bibinfo{author}{Jagersand, M.}, \bibinfo{year}{2019}.
\newblock \bibinfo{title}{Basnet: Boundary-aware salient object detection}, in: \bibinfo{booktitle}{IEEE/CVF conference on computer vision and pattern recognition}, pp. \bibinfo{pages}{7479--7489}.
\bibitem[{Ronneberger et~al.(2015)Ronneberger, Fischer and Brox}]{ronneberger2015u}
\bibinfo{author}{Ronneberger, O.}, \bibinfo{author}{Fischer, P.}, \bibinfo{author}{Brox, T.}, \bibinfo{year}{2015}.
\newblock \bibinfo{title}{U-net: Convolutional networks for biomedical image segmentation}, in: \bibinfo{booktitle}{Medical image computing and computer-assisted intervention--MICCAI 2015: 18th international conference, Munich, Germany, October 5-9, 2015, proceedings, part III 18}, \bibinfo{organization}{Springer}. pp. \bibinfo{pages}{234--241}.
\bibitem[{Salvagnini et~al.(2024)Salvagnini, Gomes, Santos, Guimar{\~a}es and Falc{\~a}o}]{salvagnini2024improving}
\bibinfo{author}{Salvagnini, F.C.R.}, \bibinfo{author}{Gomes, J.F.}, \bibinfo{author}{Santos, C.A.}, \bibinfo{author}{Guimar{\~a}es, S.J.F.}, \bibinfo{author}{Falc{\~a}o, A.X.}, \bibinfo{year}{2024}.
\newblock \bibinfo{title}{Improving flim-based salient object detection networks with cellular automata}, in: \bibinfo{booktitle}{2024 37th SIBGRAPI Conference on Graphics, Patterns and Images (SIBGRAPI)}, \bibinfo{organization}{IEEE}. pp. \bibinfo{pages}{1--6}.
\bibitem[{Sandler et~al.(2018)Sandler, Howard, Zhu, Zhmoginov and Chen}]{sandler2018mobilenetv2}
\bibinfo{author}{Sandler, M.}, \bibinfo{author}{Howard, A.}, \bibinfo{author}{Zhu, M.}, \bibinfo{author}{Zhmoginov, A.}, \bibinfo{author}{Chen, L.C.}, \bibinfo{year}{2018}.
\newblock \bibinfo{title}{Mobilenetv2: Inverted residuals and linear bottlenecks}, in: \bibinfo{booktitle}{IEEE conference on computer vision and pattern recognition}, pp. \bibinfo{pages}{4510--4520}.
\bibitem[{Schwartz et~al.(2020)Schwartz, Dodge, Smith and Etzioni}]{schwartz2020green}
\bibinfo{author}{Schwartz, R.}, \bibinfo{author}{Dodge, J.}, \bibinfo{author}{Smith, N.A.}, \bibinfo{author}{Etzioni, O.}, \bibinfo{year}{2020}.
\newblock \bibinfo{title}{Green ai}.
\newblock \bibinfo{journal}{Communications of the ACM} \bibinfo{volume}{63}, \bibinfo{pages}{54--63}.
\bibitem[{Soares et~al.(2024)Soares, Cerqueira, Guimaraes, Gomes and Falc{\~a}o}]{soares2024adaptive}
\bibinfo{author}{Soares, G.J.}, \bibinfo{author}{Cerqueira, M.A.}, \bibinfo{author}{Guimaraes, S.J.F.}, \bibinfo{author}{Gomes, J.F.}, \bibinfo{author}{Falc{\~a}o, A.X.}, \bibinfo{year}{2024}.
\newblock \bibinfo{title}{Adaptive decoders for flim-based salient object detection networks}, in: \bibinfo{booktitle}{2024 37th SIBGRAPI Conference on Graphics, Patterns and Images (SIBGRAPI)}, \bibinfo{organization}{IEEE}. pp. \bibinfo{pages}{1--6}.
\bibitem[{Sousa et~al.(2021)Sousa, Reis, Zerbini, Comba and Falc{\~a}o}]{sousa2021cnn}
\bibinfo{author}{Sousa, A.M.}, \bibinfo{author}{Reis, F.}, \bibinfo{author}{Zerbini, R.}, \bibinfo{author}{Comba, J.L.}, \bibinfo{author}{Falc{\~a}o, A.X.}, \bibinfo{year}{2021}.
\newblock \bibinfo{title}{Cnn filter learning from drawn markers for the detection of suggestive signs of covid-19 in ct images}, in: \bibinfo{booktitle}{2021 43rd Annual International Conference of the IEEE Engineering in Medicine \& Biology Society (EMBC)}, \bibinfo{organization}{IEEE}. pp. \bibinfo{pages}{3169--3172}.
\bibitem[{de~Souza et~al.(2020)de~Souza, Benato and Falc{\~a}o}]{de2020feature}
\bibinfo{author}{de~Souza, I.E.}, \bibinfo{author}{Benato, B.C.}, \bibinfo{author}{Falc{\~a}o, A.X.}, \bibinfo{year}{2020}.
\newblock \bibinfo{title}{Feature learning from image markers for object delineation}, in: \bibinfo{booktitle}{2020 33rd SIBGRAPI Conference on Graphics, Patterns and Images (SIBGRAPI)}, \bibinfo{organization}{IEEE}. pp. \bibinfo{pages}{116--123}.
\bibitem[{Tanaka et~al.(2020)Tanaka, Kunin, Yamins and Ganguli}]{tanaka2020pruning}
\bibinfo{author}{Tanaka, H.}, \bibinfo{author}{Kunin, D.}, \bibinfo{author}{Yamins, D.L.}, \bibinfo{author}{Ganguli, S.}, \bibinfo{year}{2020}.
\newblock \bibinfo{title}{Pruning neural networks without any data by iteratively conserving synaptic flow}.
\newblock \bibinfo{journal}{Advances in neural information processing systems} \bibinfo{volume}{33}, \bibinfo{pages}{6377--6389}.
\bibitem[{Wang et~al.(2021a)Wang, Joao, Falc{\~a}o, Kosinka and Telea}]{wang2021focus}
\bibinfo{author}{Wang, J.}, \bibinfo{author}{Joao, L.M.}, \bibinfo{author}{Falc{\~a}o, A.}, \bibinfo{author}{Kosinka, J.}, \bibinfo{author}{Telea, A.}, \bibinfo{year}{2021}a.
\newblock \bibinfo{title}{Focus-and-context skeleton-based image simplification using saliency maps.}, in: \bibinfo{booktitle}{VISIGRAPP (4: VISAPP)}, pp. \bibinfo{pages}{45--55}.
\bibitem[{Wang et~al.(2018)Wang, Zhang, Wang, Lu, Yang, Ruan and Borji}]{wang2018detect}
\bibinfo{author}{Wang, T.}, \bibinfo{author}{Zhang, L.}, \bibinfo{author}{Wang, S.}, \bibinfo{author}{Lu, H.}, \bibinfo{author}{Yang, G.}, \bibinfo{author}{Ruan, X.}, \bibinfo{author}{Borji, A.}, \bibinfo{year}{2018}.
\newblock \bibinfo{title}{Detect globally, refine locally: A novel approach to saliency detection}, in: \bibinfo{booktitle}{IEEE conference on computer vision and pattern recognition}, pp. \bibinfo{pages}{3127--3135}.
\bibitem[{Wang et~al.(2021b)Wang, Lai, Fu, Shen, Ling and Yang}]{wang2021salient}
\bibinfo{author}{Wang, W.}, \bibinfo{author}{Lai, Q.}, \bibinfo{author}{Fu, H.}, \bibinfo{author}{Shen, J.}, \bibinfo{author}{Ling, H.}, \bibinfo{author}{Yang, R.}, \bibinfo{year}{2021}b.
\newblock \bibinfo{title}{Salient object detection in the deep learning era: An in-depth survey}.
\newblock \bibinfo{journal}{IEEE Transactions on Pattern Analysis and Machine Intelligence} \bibinfo{volume}{44}, \bibinfo{pages}{3239--3259}.
\bibitem[{Wilcoxon(1992)}]{wilcoxon1992individual}
\bibinfo{author}{Wilcoxon, F.}, \bibinfo{year}{1992}.
\newblock \bibinfo{title}{Individual comparisons by ranking methods}, in: \bibinfo{booktitle}{Breakthroughs in statistics: Methodology and distribution}. \bibinfo{publisher}{Springer}, pp. \bibinfo{pages}{196--202}.
\bibitem[{Zhang et~al.(2017)Zhang, Wang, Lu, Wang and Yin}]{zhang2017learning}
\bibinfo{author}{Zhang, P.}, \bibinfo{author}{Wang, D.}, \bibinfo{author}{Lu, H.}, \bibinfo{author}{Wang, H.}, \bibinfo{author}{Yin, B.}, \bibinfo{year}{2017}.
\newblock \bibinfo{title}{Learning uncertain convolutional features for accurate saliency detection}, in: \bibinfo{booktitle}{IEEE International Conference on computer vision}, pp. \bibinfo{pages}{212--221}.
\bibitem[{Zhou et~al.(2024)Zhou, Lin, Yang, Lai and Xie}]{zhou2024benchmarking}
\bibinfo{author}{Zhou, H.}, \bibinfo{author}{Lin, Y.}, \bibinfo{author}{Yang, L.}, \bibinfo{author}{Lai, J.}, \bibinfo{author}{Xie, X.}, \bibinfo{year}{2024}.
\newblock \bibinfo{title}{Benchmarking deep models on salient object detection}.
\newblock \bibinfo{journal}{Pattern Recognition} \bibinfo{volume}{145}, \bibinfo{pages}{109951}.

\end{thebibliography}



\end{document}